\documentclass[journal]{new-aiaa}

 \usepackage{setspace} 
\usepackage[utf8]{inputenc}
\usepackage{textcomp}

\usepackage{graphicx}
\usepackage{amsmath}
\usepackage[version=4]{mhchem}
\usepackage{siunitx}
\usepackage{longtable,tabularx}
\usepackage{subcaption}     
\usepackage{todonotes}
\usepackage{caption}
\title{Physics-informed neural networks for spacecraft trajectory optimization with exact boundary conditions}

\author{Samuel Schönherr \footnote{Graduate Student Researcher, Institute of Theoretical and Computational Physics, Petersgasse 16/II}}
\affil{TU Graz, Graz, 8010, Austria}

\author{Günther Waxenegger-Wilfing\footnote{Professor, Institute of Computer Science, Am Hubland.}\footnote{Group Leader,  Institute of Space Propulsion, Am Langen Grund. }}
\affil{Julius Maximilian University Würzburg, Würzburg, 97074, Germany}
\affil{German Aerospace Center (DLR), Hardthausen am Kocher, 74239, Germany}

\author{Sascha Ranftl\footnote{Fellow, Division of Applied Mathematics, 182 George Street.}\footnote{Assistant Professor, School of Mechanical Engineering, 585 Purdue Mall.}\footnote{Corresponding author: sranftl@purdue.edu}}
\affil{Brown University, Providence, Rhode Island, 02912}
\affil{Purdue University, West Lafayette, Indiana, 47907}

\begin{document}

\maketitle

\begin{abstract}
We use neural networks to optimize trajectories by modeling the position as a function of time and minimizing the thrust with respect to the network parameters. This corresponds to an unsupervised training procedure in which physical laws are incorporated directly into the optimization objective. In standard physics-informed neural networks, initial and terminal states are enforced through additional weighted loss terms, resulting in a multi-objective optimization problem that is highly sensitive to the choice of loss weights and can lead to poor convergence and reduced accuracy. 
To address this challenge, we construct a simple transformation, exploiting the mathematical structure of trajectory optimization problems,  to enforce boundary conditions exactly a-priori. 
This transformation can be interpreted as a reparameterization of the solution space that embeds the boundary conditions directly into the function. This converts the original constrained optimization problem into an unconstrained one over admissible trajectories. I.e. the transformation eliminates the boundary condition penalty terms and reduces the problem to a single-objective optimization. Crucially, trajectory optimization comes with a low-dimensional boundary, which allows particularly simple and stable constructions of distance functions compared to general domains in partial differential equations. Numerical experiments demonstrate that the proposed approach meaningfully improves convergence behavior and solution accuracy.  
\end{abstract}

\newpage
\section*{Nomenclature}

{\renewcommand\arraystretch{1.0}
\noindent\begin{longtable*}{@{}l @{\quad=\quad} l@{}}

$\mathcal{N}_{\boldsymbol{\theta}}$ & neural network \\
$\boldsymbol{\theta}$ & neural-network parameters \\
$\mathbf{r}$, $\mathbf{v}$ & spacecraft position vector and velocity vector \\
$\mathbf{r}_{\boldsymbol{\theta}}$ & neural-network trajectory \\
$\rho_{\boldsymbol{\theta}}$ & transformed neural network trajectory satisfying exact boundary conditions \\
$T$ & time of flight (TOF), terminal time \\
$t$ & physical time \\
$t_0$ & initial time \\
$\tau$ & normalized time, $t/T$ \\
$\mathbf{F}$ & thrust vector \\
$R$, $V$  &  interpolation term satisfying position and velocity boundary conditions, respectively \\
$\phi(\tau)$ & distance function for velocity constraints \\
$\psi(\tau)$ & distance function for position constraints \\
$\Phi$ & gravitational potential \\

$m_s$ & spacecraft mass \\
$M_q$ & mass of gravitating body $q$ \\
$\mathbf{P}_q$ & position of gravitating body $q$ \\
$N_M$ & number of gravitating bodies \\
$N$ & number of collocation points; with subscripts BC, IC, \lq\lq data", or \lq\lq physics" \\

$\mathcal{D}$ & differential operator \\
$u$ & solution of governing differential equation \\
$\Omega$ & computational domain \\
$\partial\Omega$ & boundary of computational domain \\
$\mathcal{L}$ & loss function \\

\multicolumn{2}{@{}l}{Subscripts:} \\

BC & boundary condition \\
IC & intial condition \\
opt & optimal value \\
$0$ & initial value \\
$T$ & terminal value \\

\end{longtable*}}

\addtocounter{table}{-1}

\section{Introduction}

Machine learning has gained recent interest as a tool in astronautics and aerospace engineering (AAE), e.g. for data-driven modeling, control \cite{retagne2024adaptive, waxenegger2021reinforcement}, or design of complex systems \cite{brunton2021data}. Applications range from aerodynamic shape optimization \cite{thuerey2020deep} and turbulence modeling \cite{pioch2023turbulence} to design \cite{lam2018advances} and satellite fault diagnosis \cite{ibrahim2020machine}. Data-driven surrogates have been particularly effective in reducing the computational burden of high-fidelity numerical simulations, e.g. for model optimization, parameter estimation or uncertainty quantification \cite{ranftl2021bayesian}.  

Machine learning has also gained interest for trajectory optimization specifically. Already Ampatzis and Izzo~\cite{ampatzis2009machine} demonstrated the use of machine learning for surrogate modeling of objective functions in interplanetary trajectory design. Deep learning and/or reinforcement learning has gained particular interest recently and has been used for interplanetary missions under uncertainty~\cite{zavoli2021reinforcement}, low-thrust orbital transfer~\cite{sullivan2023designing,bonasera2023designing}, intercept guidance \cite{gaudet2020reinforcement}, planetary landing \cite{gaudet2020deep}, uncontrolled re-entry prediction \cite{salmaso2023machine} and autonomous spacecraft guidance during proximity operations~\cite{federici2021deep}. Beyond deep or reinforcement learning, many other models and algorithms have been proposed, e.g. hybrid models for collision avoidance \cite{gonzalo2021board} or  Gaussian mixture models (GMMs) \cite{fife2024probabilistic} to equip the trajectory optimization with a notion of uncertainty. Data-driven optimization has also expanded to the aviation domain, where machine learning techniques assist in designing fuel-efficient aircraft trajectories subject to operational constraints~\cite{xu2023machine}. In proximity operations, learning-based trajectory planning has been used through deep neural networks~\cite{barth2025machine,yin2020low}, incl. e.g. transformers for rendezvous scenarios~\cite{guffanti2024transformers}. 

In parallel, physics-informed neural networks (PINNs) have emerged as an interesting paradigm for solving forward and inverse problems governed by differential equations (DEs) \cite{lagaris1998artificial,dissanayake1994neural,raissi2019physics}. PINNs embed the governing laws of physics directly into the training objective of a neural network by penalizing DE residuals alongside data mismatch losses, and have since been extended into broader frameworks of physics-informed machine learning \cite{karniadakis2021physics}. Applications range from scientific computing \cite{cuomo2022scientific,faroughi2022physics} and fluid mechanics \cite{cai2021physics} to optimization tasks in engineering systems, and, in particular, also includes  PINNs for trajectory optimization \cite{seo2024solving}. 

Despite their promise, PINNs for trajectory optimization face a fundamental challenge: the need to balance multiple competing objectives. In standard formulations, the governing equations, boundary conditions, and possibly data are enforced via weighted loss terms. In particular, boundary conditions are typically imposed through penalty terms, leading to a multi-objective optimization problem whose performance is highly sensitive to the choice of loss weights. This can result in convergence issues, reduced accuracy, and significant tuning effort.

In this work, we address this challenge by proposing a simple transformation- or reparametrization principle of the PINN based on distance functions \cite{SUKUMAR2022} that enforces boundary conditions exactly \emph{a priori}, i.e. before training. For this, we identify and exploit a structural property of trajectory optimization problems that allows boundary conditions to be embedded directly into the function class, thereby converting a multi-objective learning problem into a better-conditioned single-objective optimization problem. 
This eliminates the need for boundary condition penalty terms altogether. In contrast to more general settings, we exploit the typically relatively simple geometric structure of boundary conditions in trajectory optimization specifically. We demonstrate that the imposition of exact boundary conditions via this transformation can improve both convergence behavior and solution accuracy for trajectory optimization problems.

\section{Related Work and Contribution}

\subsection{PINNs for trajectory optimization}

Already \cite{seo2024solving} applied PINNs, in their basic form, to trajectory optimization, and found that PINNs compare favorably to reinforcement learning on the same task. Since then, a number of studies utilized PINNs for trajectory optimization specifically, incl. in an aerospace engineering context. Beyond AAE, applications span intelligent vehicle tracking control~\cite{cao2025intelligent}, online trajectory regeneration for multirotors~\cite{ghanifar2025online}, and vessel trajectory prediction~\cite{alam2025physics}. These works demonstrate that PINNs can learn nonlinear kinematic and dynamic constraints directly from governing equations, but they also inherit the challenges of multi-objective formulations, where the physics-informed residuals must be balanced with trajectory boundary conditions, path constraints or possibly data. 

In light of this trade-off, recent works such as AW-EL-PINNs~\cite{li2025aw} and proportional-derivative PINNs~\cite{ghanifar2025online} develop structured approaches to constraint enforcement by tailoring the PINN architecture or loss formulation to specific control-theoretic structures.

\subsection{Constraint enforcement in PINNs}

Fundamental challenges of PINNs include stiffness and gradient pathologies \cite{wang2021understanding}, convergence sensitivity to loss scaling \cite{wang2022respecting}, and difficulties with training stability in high-dimensional or complex domains \cite{wang2022and,wang2023expert}. Various approaches have been proposed to mitigate these issues, including improved architectures, training strategies, domain decomposition, Gaussian process-based formulations \cite{Dalton2024,JONES2023109984}, and adaptive loss balancing \cite{basir2022physics}. 

A particular class of methods focuses on the exact imposition of constraints. Standard PINNs typically impose constraints approximately via penalty terms, which can lead to inaccurate constraint satisfaction, especially in stiff or ill-conditioned problems \cite{wang2022and}. This has motivated approaches that enforce constraints exactly by construction, e.g. via stochastic process formulations \cite{ranftl2022connection} or by constraining underlying fields or potentials \cite{van2025enforcing,masi2021thermodynamics}. 

Several works specifically address boundary conditions: distance function approaches \cite{SUKUMAR2022}, unified hard-constraint frameworks \cite{liu2022unified}, penalty-free neural methods \cite{SHENG2021}, Fourier feature embeddings for Neumann conditions \cite{straub2025hardconstrainingneumannboundaryconditions}, and decoupled formulations \cite{luong2025simultaneous}. Beyond neural networks, constrained Gaussian processes have been used to encode boundary conditions exactly in linear DEs \cite{Dalton2024,JONES2023109984,lange2018algorithmic,langehegermannPDEextension2023}. These approaches demonstrate that exact constraint imposition can improve accuracy, convergence, and robustness by avoiding the trade-offs inherent in penalty-based formulations \cite{rohrhofer2023data,wang2021understanding,gunantara2018review}. 

\subsection{PINNs in astronautics and aerospace applications}

In trajectory optimization within AAE, PINNs have been applied to orbital transfers, rendezvous guidance, and ascent trajectory design. Schiassi et al.~\cite{schiassi2022physics} demonstrated PINNs for optimal planar orbit transfers, while subsequent work extended to asteroid proximity maneuvers \cite{d2021physics}, low-thrust rendezvous of CubeSats \cite{das2025physics}, and orbital pursuit-evasion problems \cite{zhang2025low}. Physics-Informed Pontryagin Neural Networks (PIPNNs) incorporate optimality conditions directly \cite{d2024learning,d2025physics,benedikter2025rocket}, while hybrid methods combine PINNs with sampling-based planning \cite{drozd2024rapidly}. 

Several works also aim to improve constraint handling, including unsupervised PINNs for low-thrust transfers \cite{goldman2024unsupervised}, multi-objective planning approaches \cite{zhong2025multi}, and applications in flight dynamics and uncertainty quantification \cite{michek2024flight}. A limited number of works enforce constraints structurally, e.g. via Pontryagin-based formulations \cite{d2025physics}, trajectory-embedded transformations \cite{zhang2025low}, or constraint-regularized transfer learning \cite{badura2024regularizing}. PINNs notably have also been used successfully for orbit determination \cite{badura2025convergence}.

\subsection{Contribution}

These works confirm that boundary condition enforcement remains a central challenge in PINNs for trajectory optimization. Notably, among the existing literature, only \cite{d2024learning,d2025physics} impose certain constraints exactly using the Theory of Functional Connections, while most approaches rely on soft constraint formulations. In contrast to the Theory of Functional Connections, which constructs problem-specific basis functions with potentially high implementation complexity, our approach operates directly on standard neural network parametrizations via simple multiplicative distance functions. In contrast, \cite{badura2024regularizing} argue that exact boundary conditions are often not available in certain applications. 

In this work, we make the following contributions:

\begin{itemize}
    \item \textbf{Methodological contribution:} We propose a simple and systematic method for embedding exact kinematic constraints into PINNs for trajectory optimization \emph{a priori}, i.e. before training, via a linear transformation of the network. In contrast to \cite{zhang2025low,badura2024regularizing,d2024learning,das2025physics}, this approach avoids the need for and limitations of specialized frameworks such as the Theory of Functional Connections or Extreme Learning Machines.

    \item \textbf{Conceptual contribution:} The proposed transformation, based on distance functions, enforces initial and terminal conditions exactly by construction and eliminates the boundary condition loss term, thereby reducing the original multi-objective problem to a single-objective optimization, which is generally easier to solve. The proposed construction applies to trajectory optimization problems with kinematic constraints and extends naturally to alternative coordinate systems and conceptually to higher-order constraints, and phase-space formulations.

    \item \textbf{Computational implications:} By removing the need for balancing loss terms, the method removes sensitivity to loss weights associated with boundary conditions, which is a major source of instability in standard PINNs, and simplifies the training process.

    \item \textbf{Empirical validation:} Through a series of trajectory optimization problems of increasing complexity, including multi-body swing-by and low-thrust orbit transfer, we demonstrate improved convergence behavior, robustness, and solution quality compared to standard PINNs and competitive performance relative to a baseline defined by the classical numerical solver \textit{OpenGoddard} \cite{OpenGoddard}. The code of our PINN-implementation with exact boundary conditions is available at:  \url{https://github.com/s-schoenherr/SpacePINN}
    \end{itemize}

\section{Background}
\label{sec:headings}

We first introduce basic PINNs in Sec. \ref{sec:pinns} and then derive the appropriate loss terms for trajectory optimization problems in Sec. \ref{sec:traj-opt-with-nns}. Based on this, we will later introduce the transformation of the PINN to eliminate the boundary condition-term (BC) a priori in Sec. \ref{sec:transformed-pinn}.

\subsection{Physics-informed neural networks} \label{sec:pinns}

The central idea of physics-informed neural networks (PINNs) is to embed prior knowledge of the underlying physical process into a machine learning model~\cite{karniadakis2021physics,raissi2019physics}. This prior knowledge is typically expressed in the form of an ordinary or partial differential equation (PDE) along with corresponding boundary and initial conditions.

Let $\Omega \subset \mathbb{R}^d$ denote a spatial domain with boundary $\partial \Omega$, with time $t$ where $T > 0$ denotes the terminal time. Formally, we seek a solution $u$ in a suitable function space $\mathcal{V}$ of sufficiently smooth functions satisfying the governing equation and boundary conditions.  The governing equation is a  differential equation of the form
\begin{align} \label{eq:PDE-def}
    \mathcal{D}[u(\mathbf{x}, t)] = f(\mathbf{x}, t), \quad (\mathbf{x}, t) \in \Omega \times (0,T],
\end{align}
where $\mathcal{D}[\cdot]$ is a (possibly nonlinear) differential operator and $f$ is an inhomogeneity. This PDE is complemented by initial and boundary conditions
\begin{align}
    u(\mathbf{x}, 0) = u_0(\mathbf{x}), \quad \mathbf{x} \in \Omega,
\end{align}
\begin{align}
    u(\mathbf{x}, t) = g(\mathbf{x}, t), \quad (\mathbf{x}, t) \in \partial \Omega \times (0, T].
\end{align}

For instance, depending on the regularity requirements of the differential operator $D[\cdot]$, one may consider function spaces $\mathcal{V}$ such as $u \in C^k(\Omega \times (0,T])$, or an appropriate Sobolev space $u \in H^k(\Omega \times (0,T])$. 

A PINN approximates the solution $u$ by a neural network $u_{\mathbf{\theta}}(\mathbf{x}, t)$ that is parametrized by weights and biases summarized as $\mathbf{\theta}$. The PINN ansatz $u_{\mathbf{\theta}}$ then defines a parametrized subset $\mathcal{V}_\theta \subset \mathcal{V}$. By substituting  $u_{\mathbf{\theta}}(\mathbf{x}, t)$ into Eqs. (1)-(3), we obtain residuals each for the physical law, the initial and boundary conditions, respectively. Then, training is performed by minimizing weighted sum of these residuals, evaluated at a selected number of collocation points. Formally, the optimal weights and biases $\mathbf{\theta}_{\text{opt}}$ are given by
\begin{align} \label{eq:optimization-problem1}
    \mathbf{\theta}_{\text{opt}} = \arg \min \mathcal{L}(\mathbf{\theta})
\end{align}
where $\mathcal{L}$ is the so-called loss function. For example, the loss function can be defined as the mean squared error\footnote{Also other definitions and norms are possible} based on the $L_2$-norm as 
\begin{align} \label{eq:loss-function}
\mathcal{L}(\mathbf{\theta}) &= \mathcal{L}_{\text{data}} + \lambda_{\text{physics}} \mathcal{L}_{\text{physics}} + \lambda_{\text{BC}} \mathcal{L}_{\text{BC}} + \lambda_{\text{IC}} \mathcal{L}_{\text{IC}} \notag \\
&= \frac{1}{N_{\text{data}}} \sum_{i=1}^{N_{\text{data}}} \left| u_{\mathbf{\theta}}(\mathbf{x}_i, t_i) - u_i \right|^2 \notag \\
&\quad + \frac{\lambda_{\text{physics}}}{N_{\text{physics}}} \sum_{j=1}^{N_{\text{physics}}} \left| \mathcal{D}[u_{\mathbf{\theta}}](\mathbf{x}_j, t_j) - f(\mathbf{x}_j, t_j) \right|^2 \notag \\
&\quad + \frac{\lambda_{\text{BC}}}{N_{\text{BC}}} \sum_{m=1}^{N_{\text{BC}}} \left| u_{\mathbf{\theta}}(\mathbf{x}_m, t_m) - g(\mathbf{x}_m, t_m) \right|^2 \notag \\
&\quad + \frac{\lambda_{\text{IC}}}{N_{\text{IC}}} \sum_{n=1}^{N_{\text{IC}}} \left| u_{\mathbf{\theta}}(\mathbf{x}_n, 0) - u_0(\mathbf{x}_n) \right|^2.
\end{align}
with $N_{\text{data}}$, $N_{\text{physics}}$, $N_{\text{BC}}$, and $N_{\text{IC}}$ denote the number of data points $u_i \equiv u(\mathbf{x}_i, t_i)$ and collocation points $\mathbf{x}_k, t_k$ with predictions $ u_{\mathbf{\theta}}(\mathbf{x}_k, t_k)$ for each component $k=\{j,m,n\}$, while the coefficients $\lambda_{\text{physics}}$, $\lambda_{\text{BC}}$, and $\lambda_{\text{IC}}$ are tunable parameters that control their relative contributions during training. 
The data term (first term) enforces consistency with observed or simulated data, the PDE term (second term) penalizes violations of the governing equation, and the boundary condition (BC) and initial condition (IC) terms (third and fourth terms) enforce the boundary and initial conditions.

By minimizing $\mathcal{L}({\mathbf{\theta}})$, the network $u_{\mathbf{\theta}}(\mathbf{x},t)$ learns to approximate $u(\mathbf{x},t)$ in a way that respects both the data and the physical constraints. The collocation points for evaluating these residuals are typically composed as:
\begin{subequations}\label{eq:collocation-points}
\begin{equation} \label{eq:collocation-points_PDE}
    \{(\mathbf{x}_j, t_j)\}_{j=1}^{N_{\text{physics}}} \quad \text{in the domain} \quad \Omega \times (0, T] 
\end{equation}
\begin{equation} \label{eq:collocation-points_BC}
    \{(\mathbf{x}_m, t_m)\}_{m=1}^{N_\text{BC}} \quad \text{on the boundary} \quad \partial\Omega \times (0, T] 
\end{equation}
\begin{equation} \label{eq:collocation-points_IC}
    \{(\mathbf{x}_n, t_n)\}_{n=1}^{N_\text{IC}} \quad \text{for the initial condition}.
\end{equation}
\end{subequations}

In practice, PINNs impose the physics pointwise, i.e. only at the chosen collocation points. Consequently, residuals are typically small near these points but may increase in regions without collocation point coverage. Moreover, PINNs are sensitive to the values of $\lambda_{\text{physics}}$, $\lambda_{\text{BC}}$, and $\lambda_{\text{IC}}$. Careful selection of these weights is crucial for balanced training, and several adaptive strategies have been proposed~\cite{wang2021understanding,wang2022respecting,wang2023expert,basir2022physics}. Nevertheless, this remains an open research challenge and poses significant challenges to the application of PINNs for trajectory optimization.

In the next section, Sec. \ref{sec:traj-opt-with-nns}, we will show the naive PINN-formulation specifically for trajectory optimization problems.
We will show the numerical challenges with this plain setup in the experimental section Sec. \ref{sec:experiments}, 
and demonstrate how to fix the boundary conditions in Sec. \ref{sec:transformed-pinn}. Importantly, our approach is conceptually much simpler and computationally cheaper than adaptive tuning strategies, while simultaneously imposing the boundary conditions \emph{exactly} instead of approximately in \eqref{eq:loss-function}, by specifically exploiting the structure of trajectory optimization problems.

\subsection{Trajectory optimization with PINNs} \label{sec:traj-opt-with-nns}

We consider trajectory optimization governed by Newton’s equations, where the solution is the position $\mathbf{r}(t)$.
For a system with $N_M$ masses $M_q$ at positions $\mathbf{ P}_{q} = (x_q,y_q,z_q)$, the gravitational potential  $\Phi$ at a position $\mathbf{r} = (x,y,z)$ is given by 
\begin{align}
\Phi(\mathbf{r}) = \sum_{q=1}^{N_M} \frac{-G M_q}{\left\lVert \mathbf{r} - \mathbf{P}_{q} \right\rVert}.
\end{align}
For a spacecraft of mass $m_s$ exerting thrust $\mathbf{F}$, Newton’s equation of motion becomes
\begin{align}
m_s \frac{d^2 \mathbf{r}}{dt^2}
= -\sum_{q=1}^{N_M} G m_s M_q \frac{\mathbf{r} - \mathbf{P}_{q}}{\left\lVert \mathbf{r} - \mathbf{P}_{q} \right\rVert^3}
+ \mathbf{F}.
\end{align}
In order to use PINNs to optimize spacecraft trajectories, the physical law must be written in residual form. Introducing the  time-of-flight (TOF) $T$, we use normalized time $\tau: = t/T$ and  note that the TOF equals the terminal time if $t_0=0$. Setting a normalized spacecraft mass $m_S=1$, the thrust becomes
\begin{align} \label{eq:thrust}
\mathbf{F}(\tau) =
\frac{1}{T^2} \frac{d^2 \mathbf{r}}{d\tau^2}
+ \sum_{q=1}^{N_M} G M_q \frac{\mathbf{r}(\tau) - \mathbf{P}_{q}}{\left\lVert \mathbf{r}(\tau) - \mathbf{P}_{q} \right\rVert^3}.
\end{align}
In these units, we have $\tau \in \left[0,1\right]$ and the spacecraft takes one normalized time unit to complete any trajectory given by a boundary value problem.
We parameterize the trajectory $\mathbf{r}(\tau)$ via a neural network $\mathcal{N}_{\mathbf{\theta}}(\tau)\equiv \mathbf{r}_{\mathbf{\theta}}(\tau) $ with weights and biases ${\mathbf{\theta}}$, more precisely $\mathcal{N}_{\mathbf{\theta}} : \mathbb{R} \to \mathbb{R}^3$, $\tau \mapsto ({x}, {y}, {z})^\top$, and substitute
\begin{align}
\mathbf{r}(\tau) \approx \mathbf{r}_{\mathbf{\theta}}(\tau) \equiv \mathcal{N}_{\mathbf{\theta}}(\tau).
\end{align}
Following from this definition,  the thrust in \eqref{eq:thrust} becomes a function of the neural network and hence a function of ${\mathbf{\theta}}$. Such a neural network constitutes a parametrization of the ansatz function like any other ansatz function, with a major distinction being that closed forms are rarely available, but fast access to the gradient is provided by automatic differentiation. 
The optimization problem, for a given TOF, then is to minimize the integrated thrust magnitude subject to constraints, 
\begin{align} \label{eq:minimization-problem-def-trajectory}
\mathbf{\theta}_{\text{opt}} &=\arg \min_{\mathbf{\theta}} \mathcal{L}(\mathbf{\theta}) \notag \\&= 
\arg \min_{\mathbf{\theta}} \Bigg(\sum_{n=1}^N
\left\lVert \mathbf{F}(\tau_n) \right\rVert^2 + \text{boundary condition terms}\, \Bigg),
\end{align}
at $N$ collocation points $\tau_n \in \left[0, 1\right]$, $n=1,\cdots,N$. 
In other words, the trajectory $\mathbf{r}_{\mathbf{\theta}}(\tau)$ corresponds to the solution ansatz $ u_{\mathbf{\theta}}$ of the PINN in Sec. \ref{sec:pinns}, while the thrust \eqref{eq:thrust} corresponds to the physical law  \eqref{eq:PDE-def}, which in turn results in the optimization objective \eqref{eq:minimization-problem-def-trajectory} corresponding to \eqref{eq:optimization-problem1} and collocation points $\tau_n$ in place of \eqref{eq:collocation-points_PDE} in Sec. \ref{sec:pinns}.

The initial and terminal conditions of the trajectory are realized   by a starting position vector $\mathbf{r}(t_0)$ at time-zero $t_0$ and a terminal position vector $\mathbf{r}(T)$ at terminal time $T$, as well as possible constraints on the time-derivatives of the position vectors at trajectory start and trajectory end, i.e. $\dot {\mathbf{r}}(t_0)$, $\dot{\mathbf{r}}(t_T)$, $\ddot {\mathbf{r}}(t_0)$, $\ddot{\mathbf{r}}(T)$, etc.\.. 
In the naive PINN-formulation, the boundary conditions of a trajectory, i.e. here the initial and terminal conditions, are enforced through additional loss terms at collocation points corresponding to the initial and boundary condition in \eqref{eq:collocation-points_BC}-\eqref{eq:collocation-points_IC}.
Therefore, the full loss function in  \eqref{eq:minimization-problem-def-trajectory} writes as 
\begin{equation} \label{eq:def-pinn-loss-with-BCs-for-trajectories}
    \mathcal{L}(\mathbf{\theta)}= \sum_{n=1}^N
\left\lVert \mathbf{F}(\tau_n) \right\rVert^2\ + \lambda_1   
\left\lVert \mathbf{r}(t_0) - \mathbf{r}_{\mathbf{\theta}}(t_0) \right\rVert^2  + \lambda_2   
\left\lVert \mathbf{r}(T) - \mathbf{r}_{\mathbf{\theta}}(T) \right\rVert^2 + \lambda_3   
\left\lVert \dot{\mathbf{r}}(t_0) - \dot{\mathbf{r}}_{\mathbf{\theta}}(t_0) \right\rVert^2 + \lambda_4   
\left\lVert \dot{\mathbf{r}}(T) - \dot{\mathbf{r}}_{\mathbf{\theta}}(T) \right\rVert^2 + \cdots.
\end{equation} 
Then the optimized trajectory is given by the solution to \eqref{eq:minimization-problem-def-trajectory} via the PINN prediction at optimal weights and biases, $\mathbf{r}_{\text{opt}}(\tau)\equiv \mathbf{r}_{\mathbf{\theta}_\text{opt}}(\tau)$, i.e. setting ${\mathbf{\theta}} = {\mathbf{\theta}}_{\text{opt}}$ in the ansatz function. This procedure corresponds to training an unsupervised PINN as introduced before in Sec. \ref{sec:pinns} where the physics-loss $\mathcal{L}_{\text{physics}}$ is defined as the integrated thrust magnitude.

Conveniently, the formulation with normalized time allows us to treat the TOF $T$ explicitly as an optimization parameter that can be backpropagated and learned in the same way as the weights and biases $\mathbf{\theta}$, formally
\begin{align} \label{eq:minimization-problem-def-trajectory-terminal-time}
(\mathbf{\theta}_{\text{opt}}, T_{\text{opt}}) =
\arg \min_{\mathbf{\theta}, T} \mathcal{L}(\mathbf{\theta}).
\end{align}
In the implementation, this is as simple as adding $T$ to the list of optimizable parameters in addition to $\mathbf{\theta}$. In addition, 
Eq. \eqref{eq:minimization-problem-def-trajectory-terminal-time} reveals a major advantage of PINNs over classical solvers: the quantity of interest becomes auto-differentiable with respect to the physical parameters, be it TOF or other optimizable parameters. This in turn enables \textit{efficient} gradient-based approaches to inverse problems. Further, we do not need a fixed discretization of the domain $\Omega \times (0,T]$ at inference time.

However, it is clear that even adaptive strategies face an increasingly difficult problem of tuning the $\lambda$'s in the formulation of \eqref{eq:def-pinn-loss-with-BCs-for-trajectories}. 
Our experiments will demonstrate in Sec. \ref{sec:experiments} that this naive PINN formulation leads to low accuracy in satisfying the boundary conditions, slow convergence and sometimes even prevents us from convergence to a satisfying solution at all. In the next section, we introduce a simple transformation of the PINN to eliminate the boundary conditions in the loss function, thereby guaranteeing exact fulfillment of boundary conditions and eliminating convergence issues, and tuning effort, associated with the parameters $\lambda$ in \eqref{eq:def-pinn-loss-with-BCs-for-trajectories}.

\section{Exact boundary conditions by PINN-transformation}
\label{sec:transformed-pinn}

We now introduce a transformed trajectory representation that satisfies the boundary conditions (BCs) exactly.
While for the naive PINN formulation we had $\mathbf{r}_{\mathbf{\theta}}(\tau)\equiv \mathcal{N}_{\mathbf{\theta}}(\tau)$, we define the transformed PINN $\rho_\mathbf{\theta}$ with exact boundary conditions as a function of $\mathbf{r}_{\mathbf{\theta}}$ as 
\begin{align} \label{eq:rho-def}
\mathbf{\rho}_\mathbf{\theta}(\tau) :=
\mathbf{R}(\tau)
+ \psi(\tau) \mathbf{r}_{\mathbf{\theta}}(\tau)
+ \phi(\tau) \Big( \mathbf{V}(\tau) - \dot{\mathbf{R}}(\tau) \Big),
\end{align}
with its time-derivative
\begin{align} \label{eq:rho-derivative}
\dot{\mathbf{\rho}}_\mathbf{\theta}(\tau) &=
\dot{\mathbf{R}}(\tau)
+ \dot{\psi}(\tau) \mathbf{r}_{\mathbf{\theta}}(\tau)
+ \psi(\tau) \dot{\mathbf{r}}_{\mathbf{\theta}}(\tau)  + \phi(\tau) \Big( \dot{\mathbf{V}}(\tau) - \ddot{\mathbf{R}}(\tau) \Big)
+ \dot{\phi}(\tau) \Big( \mathbf{V}(\tau) - \dot{\mathbf{R}}(\tau) \Big),
\end{align}
where $\phi$ and $\psi$ are so-called distance functions \cite{SUKUMAR2022} constructed such that they guarantee the boundary conditions to be fulfilled by $\rho_\mathbf{\theta}$ and $\dot{\rho}_\mathbf{\theta}$ at the boundary for all parameter values of $\mathbf{\theta}$ in conjunction with following choice:
\begin{align} \label{eq:bc-enforcement-part1}
\mathbf{R}(\tau) &= \tau (\mathbf{r}_T - \mathbf{r}_0) + \mathbf{r}_0, \\
\mathbf{V}(\tau) &= \tau (\mathbf{v}_T - \mathbf{v}_0) + \mathbf{v}_0,\label{eq:bc-enforcement-part2}
\end{align}
where $\mathbf{r}_T = \mathbf{r}(T)$, $ \mathbf{r}_0=\mathbf{r}(t_0)$, $\mathbf{v}_T = \dot{\mathbf{r}}(T)$, $\mathbf{v}_0=\dot{\mathbf{r}}(t_0)$ is defined by the boundary conditions\footnote{Note the slight abuse of notation in that we write $\mathbf{r}(t)$ and $\mathbf{r}(\tau)$ both with the same symbol $\mathbf{r}$ albeit the expression slightly differs depending on whether the argument is time or normalized time. We here use un-normalized time for clarity of argumentation regarding the requirements for $\phi$ and $\psi$, and avoid overloading notation to distinguish $\mathbf{r}(t)$, $\psi(t)$ from their normalized time-counterparts e.g. by $\tilde{\mathbf{r}}(\tau)$ and $\tilde{\psi}(\tau)$. Note that this normalization must be anticipated in the boundary conditions.}.
Then, for the transformed PINN to satisfy the boundary conditions $\forall \mathbf{\theta}$, we need to construct (i) $\psi$ such that $\psi(t_0) = 0$, $\psi(T) = 0$, $\dot{\psi}(t_0) = 0$ and $\dot{\psi}(T) = 0$, 
as well as (ii) $\phi$ such that $\phi(t_0) = 0$, $\phi(T) = 0$, $\dot{\phi}(t_0) = 1$ and $\dot{\phi}(T) = 1$.  
Setting $t_0=0$ and recalling normalized time $\tau:=t/T, ~\tau \in [0,1]$, one possible solution is 
\begin{align} \label{eq:bc-enforcement-part3}
\psi(\tau) &= \tau^2 (1 - \tau)^2, \\
\phi(\tau) &= 2 \tau^3 - 3 \tau^2 + \tau. \label{eq:bc-enforcement-part4}
\end{align}
We may then solve the trajectory optimization problem with exact boundary conditions by substituting $\mathbf{\rho}_{\mathbf{\theta}}$ for $\mathbf{r}_{\mathbf{\theta}}$ in Eq. \ref{eq:thrust} and Eq.~\eqref{eq:minimization-problem-def-trajectory}, respectively. 
In this construction, the transformed PINN is always and a priori consistent with the boundary conditions and the velocity $\mathbf{v}(t) \equiv \dot{\mathbf{\rho}}(t)$, which is easily verified by the substitution of \eqref{eq:bc-enforcement-part1}-\eqref{eq:bc-enforcement-part4} into \eqref{eq:rho-def} and \eqref{eq:rho-derivative}. This consistency is not fulfilled in the ordinary PINN, nor in other constructions such as two neural networks learning position and velocity independently, respectively, nor by a single neural network with multiple outputs learning position and velocity jointly.
 
\subsection{Implementation remarks}
We construct a neural network $\mathcal{N}_{\mathbf{\theta}} : \mathbb{R} \to \mathbb{R}^3$, $\tau \mapsto ({x}, {y}, {z})^\top$, and define the transformed trajectory $\mathbf{\rho}_{\mathbf{\theta}}(\tau)$ via the transformation above. I.e. both $\mathbf{r}_{\mathbf{\theta}}$ and $\mathbf{\rho}_{\mathbf{\theta}}$ take time as input and yield a three-dimensional vector as output. By construction, $\mathbf{\rho}_{\mathbf{\theta}}(\tau)$ satisfies the boundary conditions for all $\mathbf{\theta}$ and remains fully differentiable with respect to both $\tau$ and $\mathbf{\theta}$.
Training is then achieved by solving the optimization problem with either fixed TOF Eq.~\eqref{eq:minimization-problem-def-trajectory} or optimizable TOF Eq.~\eqref{eq:minimization-problem-def-trajectory-terminal-time}, and the argument of free terminal time from the PINN in Sec. \ref{sec:pinns} carries over to the transformed PINN. 
The explicit expression for $\dot{\mathbf{\rho}}_{\mathbf{\theta}}(\tau)$ is not required in implementation, since automatic differentiation provides all necessary derivatives. In other words, the derivative expression \eqref{eq:rho-derivative} is only used to derive suitable conditions for $\phi(\tau)$ and $\psi(\tau)$ to enforce the boundary conditions.

\subsection{Generalizations}
Regarding construction of the transformation, it may be advantageous to introduce an alternative distance function $\chi(\tau)$ and consider a modified construction of the form $\chi(\tau)\,\mathbf{R}(\tau)$. This can be used to eliminate $\dot{\mathbf{R}}(\tau)$ at the boundary $\tau \in \{0,1\}$ and replace the term $\big( \mathbf{V}(\tau) - \dot{\mathbf{R}}(\tau) \big)$ by $\mathbf{R}(\tau)$. 
Moreover, the distance functions $\phi(\tau)$, $\psi(\tau)$, or $\chi(\tau)$ may themselves contain learnable parameters. The proposed construction also naturally extends beyond kinematic constraints, for instance to (i) alternative coordinate systems such as cylindrical or spherical coordinates, (ii) higher-order constraints by extending the construction to $\ddot{\mathbf{r}}$, or (iii) constraints in phase space. In particular, parametrizing higher-order distance functions may enable the learning of maneuvers with short impulses or acceleration limits and shall be the subject of future work.

\subsection{Interpretation as constraint embedding of a Rayleigh-type variational problem}
The proposed transformation can be interpreted as a reparameterization of the solution space that embeds the boundary conditions directly into the function class. While for the ordinary PINN without exact BCs (i.e. only soft BCs) we have $\mathbf{r}_{\mathbf{\theta}}\in \mathcal{V}_{\mathbf{\theta}}$ with $\mathcal{V}_{\mathbf{\theta}} \subset \mathcal{V}$ (cf. Sec.\ref{sec:pinns} and recall our definition $u_{\mathbf{\theta}} \equiv \mathbf{r}_{\mathbf{\theta}}$), the transformed PINN with exact BCs is  $\mathbf{\rho}_{\mathbf{\theta}}\in \mathcal{V}_{\mathrm{BC}}$ with $\mathcal{V}_{\mathrm{BC}} \subset \mathcal{V_\mathbf{\theta}}$.
Instead of enforcing constraints through penalty terms, which leads to a multi-objective optimization problem, the transformation restricts the function space generated by the neural network to functions that satisfy the boundary conditions by construction. As a result, the original constrained optimization problem is converted into an unconstrained optimization over admissible trajectories. This perspective explains both the improved convergence behavior and the reduced sensitivity to hyperparameters observed in the numerical experiments. Lastly, the resulting optimization problem corresponds to a continuous-thrust generalization of a Lambert boundary value problem and to a Rayleigh-type variational problem in which the integrated control effort is minimized over the space of admissible trajectories.

\section{Numerical Experiments}
\label{sec:experiments}
 
We demonstrate five numerical experiments of increasing complexity: (A) a swing by trajectory for a spacecraft passing three celestial bodies in 2D,  (B) a swing by trajectory for a spacecraft passing three celestial bodies in 3D, (C) a circular orbital transfer in polar coordinates with fixed terminal angle, (D) a circular orbital transfer in polar coordinates with free terminal angle, and (E) hold-point rendezvous maneuver in low earth orbit. In all experiments, we compare our transformed PINN with exact boundary conditions (BCs) to a baseline obtained with the classical numerical solver OpenGoddard \cite{OpenGoddard}. Where possible, we also compare our transformed PINN with exact boundary conditions to the naive PINN without exact boundary conditions (i.e. soft BCs). However, in examples (C)-(E) the naive PINNs failed to converge to a satisfactory solution, suggesting that our method from Sec. \ref{sec:transformed-pinn} enables more difficult problems that are beyond reach with the naive PINN. In all cases, we compare to the \lq\lq best\rq\rq~ naive PINN, meaning that the PINN with optimized weighting parameters, i.e. the $\lambda$'s in \eqref{eq:loss-function}, is selected. Supplementary material on the grid search and an illustration of the computational challenges to select the best $\lambda$'s is provided in Appendix \ref{app:hyperparameter-tuning}. 
In all experiments, we report the velocity increment defined via the integrated thrust magnitude as
\begin{align} \label{eq:deltaV}
    \Delta V = \int_0^T \frac{
    \lvert\lvert\mathbf{F}(t)\rvert\rvert}{m(t)} \, dt
\end{align}
where for the sake of simplicity we set $m(t)\equiv 1$. The method can in principle be extended to non-constant masses. The integral was computed using 1000 equi-spaced collocation points, i.e. ten times as many as during training. Note that the thrust-term in \eqref{eq:minimization-problem-def-trajectory} is a discrete approximation closely related to $\Delta V$ in \eqref{eq:deltaV}.
  
For all neural networks, we used 3 layers with 50 densely connected neurons each. The input was the normalized time, and the output was a position in 2D- or 3D-space, respectively. For the examples of Sec. \ref{sec:example1}-\ref{sec:example2}, this output was in cartesian coordinates, whilst examples of Sec. \ref{sec:example3}-\ref{sec:example5} were in polar coordinates. For training, ADAM and LBFGS optimizers were used consecutively. The trajectory was considered converged after the relative change between two training iterations is smaller than $\varepsilon$, with  $\varepsilon=10^{-6}$ in the experiments of Sec. \ref{sec:example1}-\ref{sec:example2} and $\varepsilon = 10^{-5}$ in the experiments of Sec. \ref{sec:example3}-\ref{sec:example5}. The number of ADAM iterations was set individually for all experiments while the maximum number of iterations for the LBFGS optimizer was set to $10^5$ in the experiments of Sec. \ref{sec:example1}-\ref{sec:example2}. No LBFGS was necessary in the experiments of Sec. \ref{sec:example3}-\ref{sec:example5}. The loss term was computed on 100 linearly equi-spaced collocation points for normalized time.

Unless stated otherwise, examples reporting \textit{normalized units} (Sec. \ref{sec:example1}, Sec, \ref{sec:example2}) employ dimensionless variables, where time is normalized by the TOF, the spacecraft mass is set to unity, and spatial coordinates are scaled such that the computational domain spans a unit square. The resulting $\Delta V$ for all examples across all methods is summarized in Sec. \ref{sec:summary}. All examples report figures of the trajectories, training curves, required force magnitude (RFM) and gravity experienced by the spacecraft and thrust magnitude along the trajectory.
The computational time for PINN training and the classical solver is comparable and on the order of minutes for the demonstrated examples.

All codes are implemented in PyTorch and available at \url{https://github.com/s-schoenherr/SpacePINN}.
We defer a parameter study with respect to fixed and free TOF to Appendix \ref{app:C}.

\subsection{Swing-by with three celestial bodies in 2D} \label{sec:example1}
We seek to optimize a swing-by trajectory for a spacecraft in a given gravitational potential and given boundary conditions. This problem was originally posed by Seo  \cite{seo2024solving}. We will start with two spatial dimensions and then extend to three dimensions. The gravitational potential is defined by three celestial body coordinates and masses in the $(x,y)$ plane in Tab. \ref{tab:astronomic-objects}.

\begin{table}[h!]
    \centering
    \caption{Positions and gravitational masses $GM_i$ of celestial bodies}
    \label{tab:astronomic-objects}
    \begin{tabular}{ccc}

    \hline\hline
        \# &$(x,y,z) \left[ m\right]$ & $GM_i \left[ m^3s^{-2}\right]$ \\ \hline
        1&$(-0.5, -1.0,0)$ & 0.5 \\ 
        2&$(-0.2,+0.4,0)$ & 1.0 \\ 
        3&$(+0.8,+0.3,0)$ & 0.5 \\ \hline\hline
    \end{tabular}
\end{table}

In this first example, our neural network will map $\mathcal{N}_{\mathbf{\theta}} : \mathbb{R} \to \mathbb{R}^2$, $\tau \mapsto ({x}, {y})^\top$, omitting the $z$-coordinate. We will later extend this to three dimensions. We do this to rule out solutions that deviate from the $(x,y)$-plane and isolate a direct comparison in the first experiment, whilst such solutions are permitted in the second experiment.

\begin{figure}[htbp]
    \centering
    \begin{subfigure}[b]{0.49\linewidth}
        \centering
        \includegraphics[width=\linewidth]{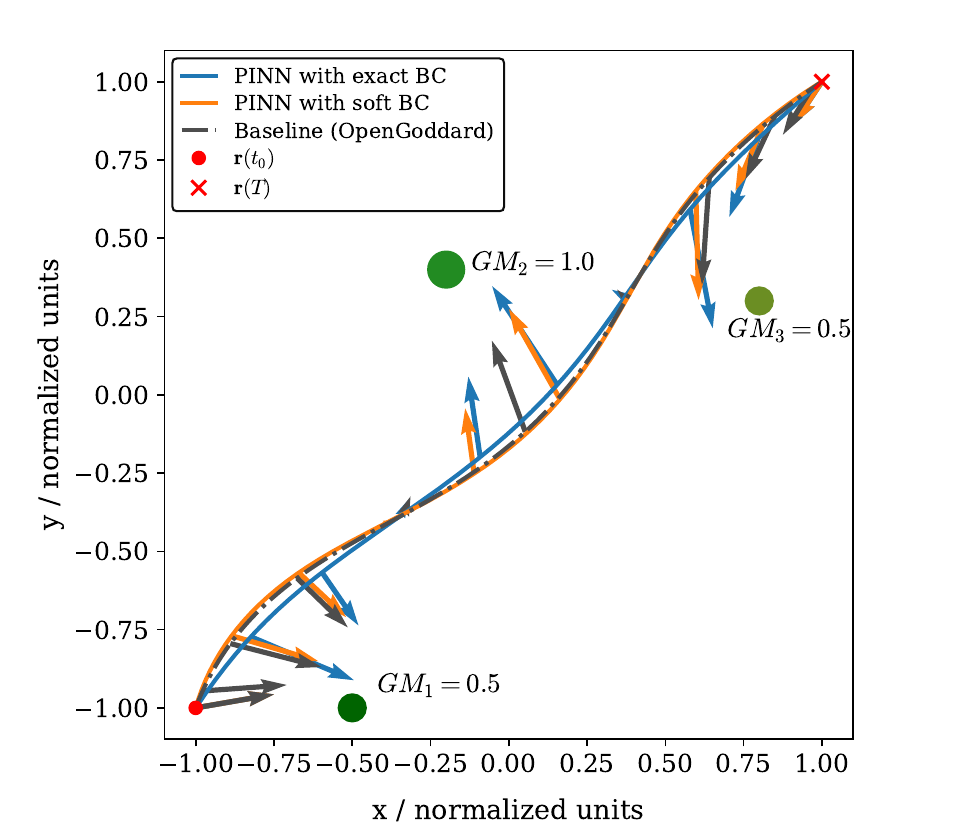}
        \caption{Trajectories. Green disks are celestial bodies.}
        \label{fig:PV2d_traj}
    \end{subfigure}
    \begin{subfigure}[b]{0.49\linewidth}
        \centering
        \includegraphics[width=\linewidth]{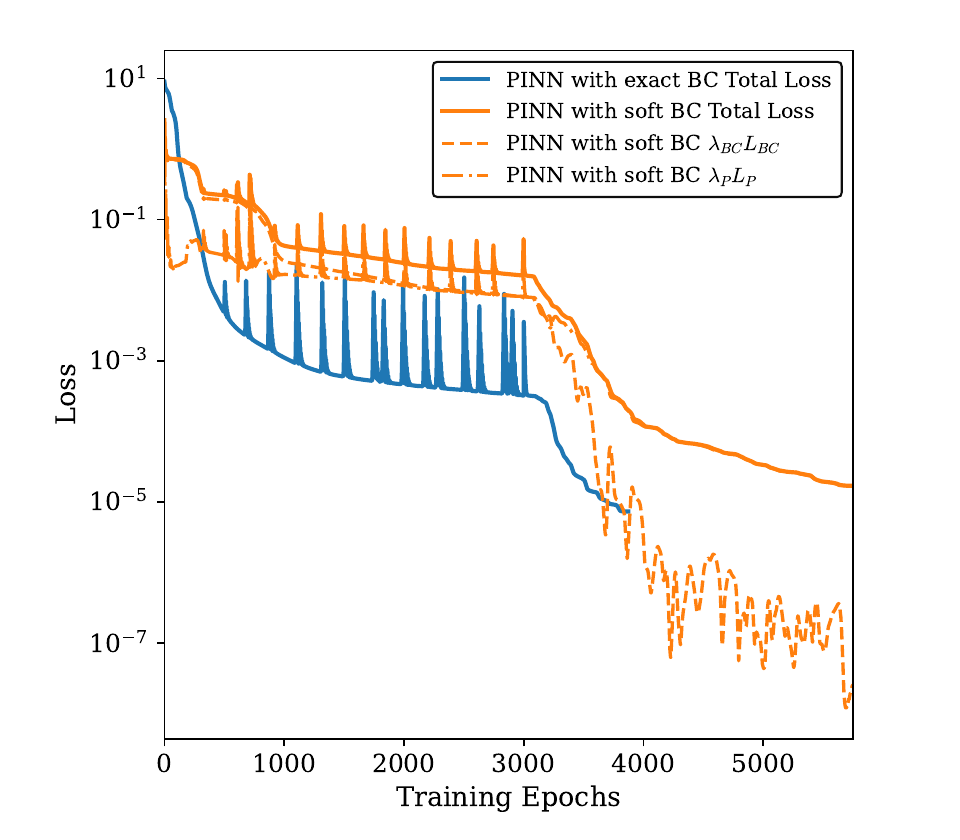}
        \caption{Training curve.}
        \label{fig:PV2d_loss}
    \end{subfigure}
    \begin{subfigure}[b]{0.49\linewidth}
        \centering
        \includegraphics[width=\linewidth]{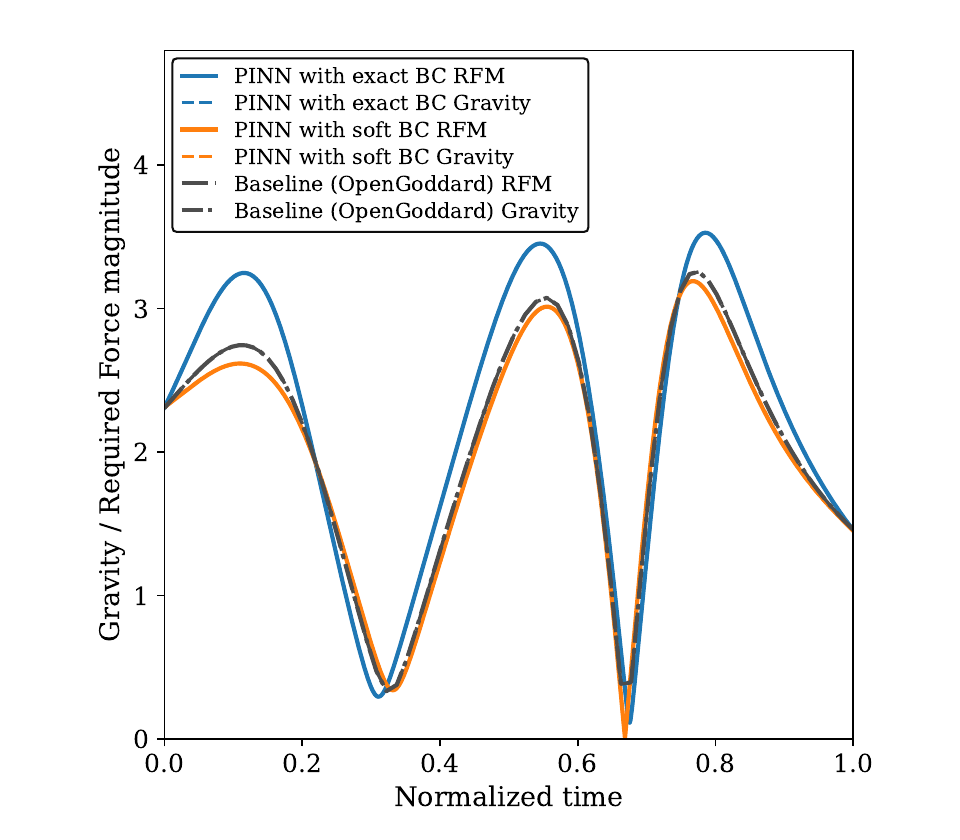}
        \caption{Gravity and required force magnitude (RFM).}
        \label{fig:PV2d_gravity}
    \end{subfigure}
    \begin{subfigure}[b]{0.49\linewidth}
        \centering        \includegraphics[width=\linewidth]{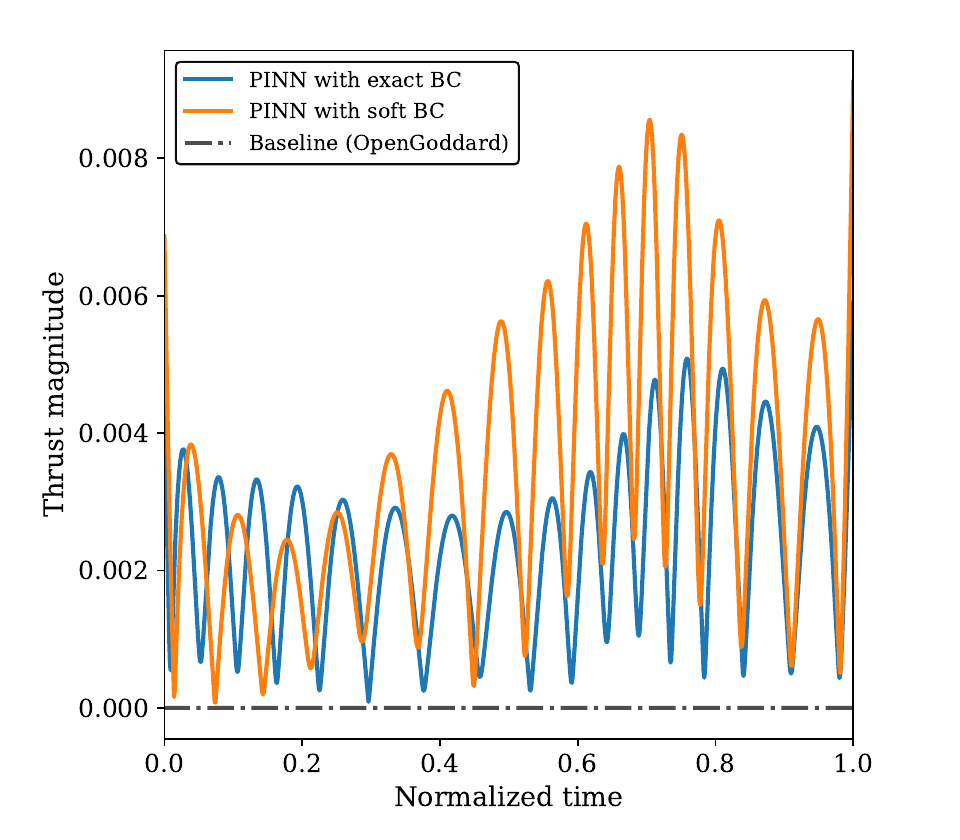}
        \caption{Thrust magnitude.}
        \label{fig:PV2d_thrust}
    \end{subfigure}
\caption{Swing-by trajectory in 2D. Comparison of the naive PINN, our method (exact BC), and the baseline. All units normalized. Arrows scaled for visualization.}
    \label{fig:PV2d}
\end{figure}

The boundary conditions are $\textbf{r}(t_0)=(-1,-1)$ and $\textbf{r}(T)=(1,1)$, respectively. The initial and terminal state velocities remain free, i.e. we set $\phi(t) \equiv 0$ in \eqref{eq:rho-def}.
The TOF is optimizable with an initial guess of $T=1$. We found local optima for the TOF close to the initial value, however hyperparameter tuning is more sensitive with optimizable TOF.
The results for both the best naive PINN ($\lambda_{BC}=0.324$) and the transformed PINN with exact boundary conditions are shown in Fig. \ref{fig:PV2d_traj}. Whilst both trajectories show reasonable qualitative agreement with the classical solver's baseline, the PINNs cannot compete in terms of integrated thrust magnitude $\Delta V$ (cf. Tab. \ref{tab:comparison_methods}). While the transformed PINN substantially improves convergence behavior and exact satisfaction of boundary conditions compared to the naive PINN, the classical solver remains superior in terms of integrated thrust optimality. We attribute this to the fact that the PINN solves the variational problem only approximately through residual minimization at finitely many collocation points and through a highly nonconvex neural-network parametrization.
The solution trajectory (top left panel) of the naive PINN is qualitatively similar to the transformed PINN with exact boundary conditions. However,  the naive PINN deviates substantially from the boundary conditions and converges significantly slower to an optimum with larger physics loss (top right panel).  The convergence curve also shows that the naive PINN stalls at a residual error for the boundary conditions on the order of $10^{-4}$. Lastly, the acting gravity and thrust magnitude (bottom panels) differ substantially quantitatively between the two models. In particular, transformed PINN with exact boundary conditions achieves consistently lower required thrust compared to the naive PINN, see also summary metrics and statistics in Sec. \ref{sec:summary}.

We found that such differences in the convergence curves persist throughout the experiments, and we offer a simple explanation for the phenomenon. While the naive PINN needs to balance the satisfaction of the boundary conditions against minimizing the thrust, the transformed PINN need not make any such balance and solely needs to minimize the thrust.

\subsection{Swing-by with three celestial bodies in 3D}\label{sec:example2}
We extend the previous 2D experiment in Sec. \ref{sec:example1} to a plane in three dimensions. The celestial body coordinates as well as the boundary conditions remain the same as in Sec. \ref{sec:example1} within the $(x,y)$-plane, i.e. the $z$-coordinates of the boundary conditions and celestial body coordinates set to zero. The output of the PINNs is now a position in 3D, i.e.  $\mathcal{N}_{\mathbf{\theta}} : \mathbb{R} \to \mathbb{R}^3$, $\tau \mapsto ({x}, {y}, {z})^\top$.  

\begin{figure}[htbp]
    \centering
    \begin{subfigure}[b]{0.49\linewidth}
        \centering
        \includegraphics[width=\linewidth]{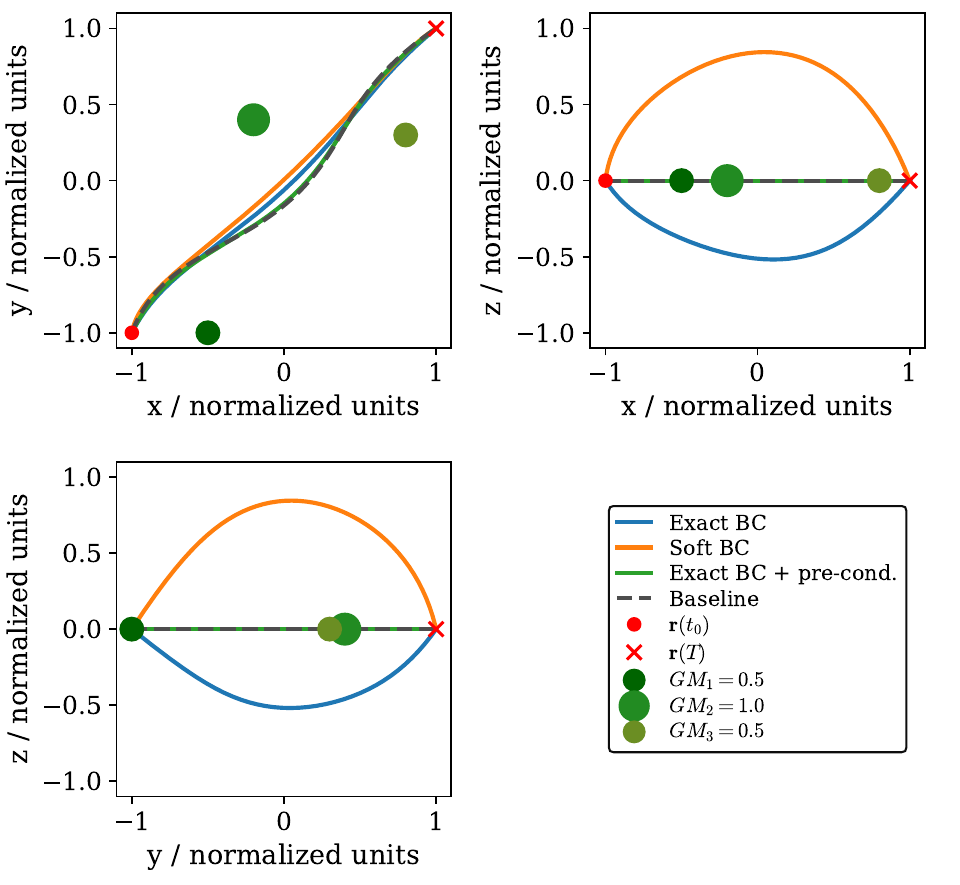}
        \caption{Trajectories. Green disks are celestial bodies.}
        \label{fig:PV3d_traj2d}
    \end{subfigure}
        \begin{subfigure}[b]{0.49\linewidth}
        \centering
        \includegraphics[width=\linewidth]{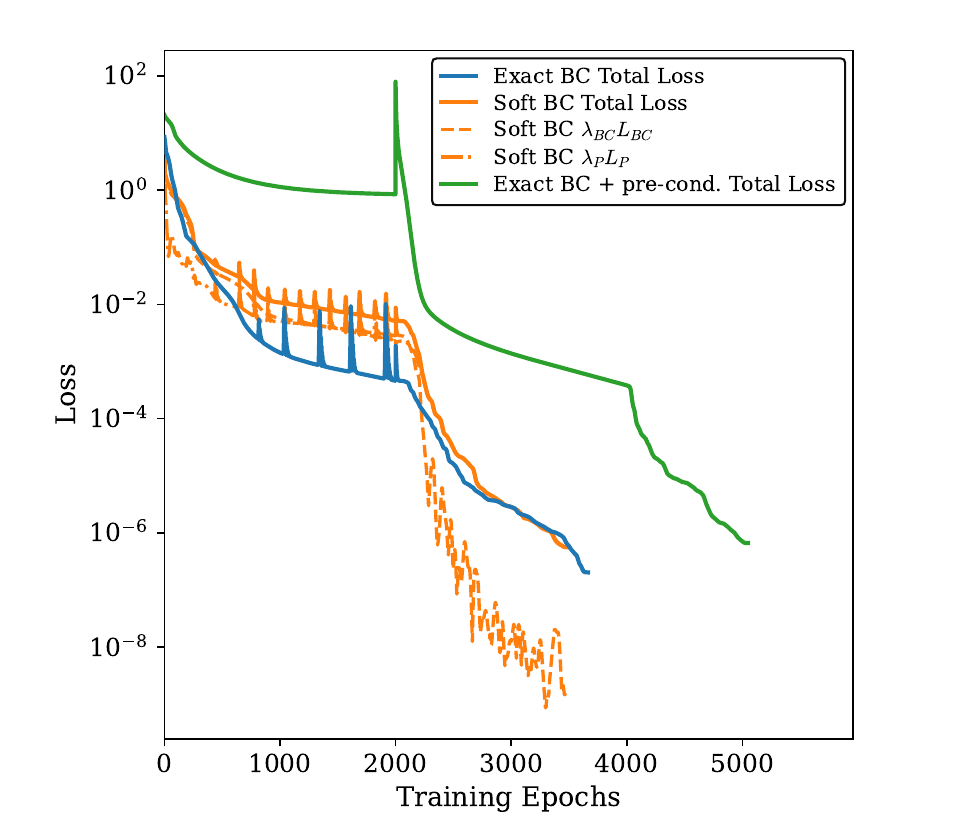}
        \caption{Training curve.}
        \label{fig:PV3d_loss}
    \end{subfigure}
    
    \begin{subfigure}[b]{0.49\linewidth}
        \centering
        \includegraphics[width=\linewidth]{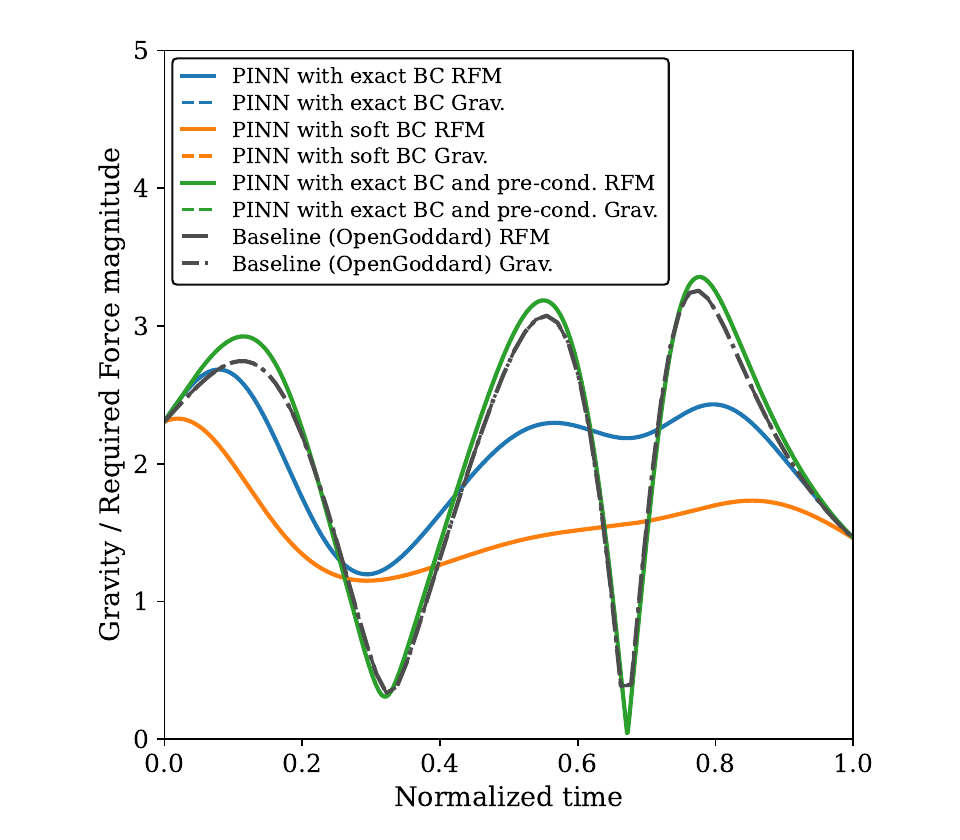}
        \caption{Gravity and required force magnitude (RFM).}
        \label{fig:PV3d_gravity}
    \end{subfigure}
    \begin{subfigure}[b]{0.49\linewidth}
        \centering
        \includegraphics[width=\linewidth]{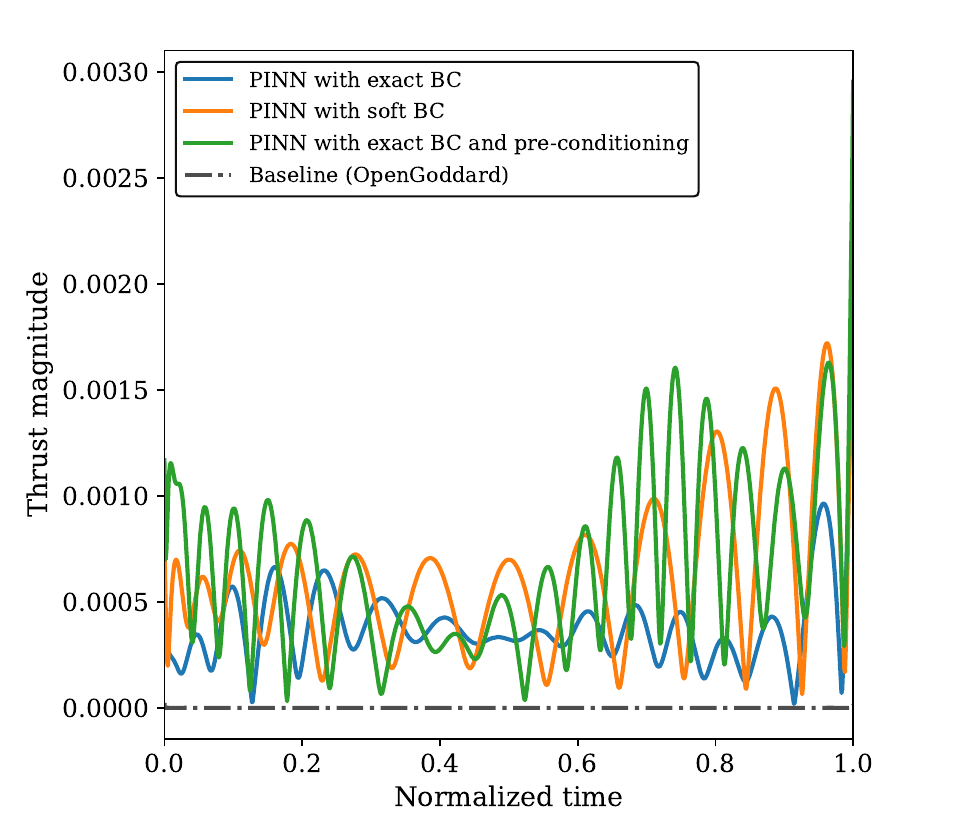}
        \caption{Thrust magnitude.}
        \label{fig:PV3d_thrust}
    \end{subfigure}

    \caption{Swing-by trajectory in 3D. Comparison of the naive PINN, our method (exact BC) with and without preconditioning, and the baseline. All units normalized. Arrows scaled for visualization.}
    \label{fig:PV3d}
\end{figure}

The results for both  the optimal naive PINN (with $\lambda_{BC}=0.421$) and the transformed PINN with exact boundary conditions are shown in Fig. \ref{fig:PV3d}. The naive PINN converges to a solution trajectory (top left panel set) that leaves the $(x,y)$-plane and circumvents the gravitational centers altogether \footnote{We confirmed that this is a physically meaningful solution as follows: The naive PINNs solution was used as an initial guess for a separate run with the numerical solver OpenGoddard, which then in turn would converge to a qualitatively similar solution. In contrast, OpenGoddard initialized with a straight line between the initial and terminal position converged to the baseline trajectory shown in  Fig. \ref{fig:PV3d}.}.

In contrast, the solution of the transformed PINN with exact boundary conditions still finds an in-plane trajectory qualitatively similar to the baseline. 
This however demanded a more sophisticated initialization or pre-conditioning scheme as follows: First, we start by initializing a transformed PINN with auxiliary boundary conditions for initial and final state velocities $\mathbf{v}_0=\mathbf{v}_T=[1,1,0]$, in addition to the initial and terminal positions, in order to prime the PINN for an in-plane trajectory. Second, after pre-training for 2000 iterations with ADAM, we find a sub-optimal trajectory in the $(x,y)$-plane. Third, the auxiliary boundary conditions are relaxed again, i.e. training resumes with the pre-conditioned PINN, pre-optimized TOF as well as now free (optimizable) initial and terminal velocities. 
We found that the transformed PINN with exact boundary conditions converges to a qualitatively similar solution as the optimal naive PINN unless this pre-conditioning step is applied. We interpret the pre-conditioning step as an additional, potentially symmetry breaking implicit biases. Importantly, this pre-conditioning step is not available in the naive PINN formulation without escalation of hyperparameter tuning for lack of a direct mechanism to impose boundary conditions exactly. 

Regarding convergence and thrust as a function of time, we have similar findings as in the 2D example, with the notable exception that it is indeed more favorable to escape the $(x,y)$-plane for a minimal thrust swing-by, as is confirmed by the convergence curve (top right panel).

\subsection{Circular Orbit Transfer with Fixed Terminal Angle} \label{sec:example3}
Next we consider an orbit transfer problem from a circular Low Earth Orbit (LEO) to a higher circular Medium Earth Orbit (MEO). For this, our method from Sec. \ref{sec:transformed-pinn} can be extended straightforward to polar coordinates with polar unit vectors $\mathbf{e}_r$ and $\mathbf{e}_\alpha$ as follows.
\begin{align}
    \mathbf{r} &= r \mathbf{e}_r \notag\\
    \dot{\mathbf{r}} &= \dot{r} \mathbf{e}_r + r \dot{\alpha} \mathbf{e}_\alpha  \notag \\
    \ddot{\mathbf{r}} &= (\ddot{r} - r\dot{\alpha}^2) \mathbf{e}_r + (2\dot{r}\dot{\alpha} + r\ddot{\alpha}) \mathbf{e}_\alpha \label{eq:acceleration_polar}
\end{align}
with $\alpha$ being the azimuthal angle coordinate in radians and $r$ the norm of $\mathbf{r}$.
We may substitute $\mathbf{r}(\tau)$ for a neural network $\mathbf{r}_{\mathbf{\theta}}: \mathcal{N}_{\mathbf{\theta}}(\tau) \to \big(r(\tau),\alpha(\tau)\big)$ into \eqref{eq:minimization-problem-def-trajectory} to obtain a PINN that outputs polar coordinates $(r,\alpha)$ as a function of normalized time. Note that one must use the acceleration in polar coordinates \eqref{eq:acceleration_polar} for the thrust vector \eqref{eq:thrust}. The transformation of the PINN to satisfy boundary conditions is then analogous to Sec. \ref{sec:transformed-pinn}, however for this type of problem the boundary conditions typically are slightly different.
We may now consider boundary conditions
\begin{align}
    \mathbf{r}_0 &= \mathbf{r}(t_0) =  \big(r_0 , \alpha_0\big)^T \notag \\
     \mathbf{r}_T &= \mathbf{r}(T) = \big(r_T, \alpha_T\big)^T.
\end{align}
with $r_0 = R_{\text{LEO}}=6871~\text{km}$, corresponding to 500 km above ground, and $r_T = R_{\text{MEO}}= 8371~\text{km}$, corresponding to 2000 km above ground, and a fixed radian phases of $\alpha_0=0$ and $\alpha_T=\pi$.
In orbital transfer problems, we may typically desire to constrain the orbits through $r(t_0)$ and $r(T)$ as well as radial velocity $\dot{r}(T)=0$ but refrain from constraining the angular phases $\alpha(t_0)$, $\alpha(T)$ or tangential velocity $\dot{\alpha}(\tau)$. For this experiment, we do constrain the phases as defined above in order to obtain a solution that is comparable to an analytic solution of a Hohmann-transfer. We will relax this constraint in another experiment later in Sec. \ref{sec:example4}.
The initial and terminal orbital velocities of the LEO and MEO were constrained to 
$\dot{\alpha}(t_0) = \frac{V_{\mathrm{LEO}} }{R_{\mathrm{LEO}} }$ with $V_{\mathrm{LEO}}  =\sqrt{\frac{GM_{\mathrm{Earth}}}{R_{\mathrm{LEO}}}} = 7.61~\mathrm{km/s}$ and 
$\dot{\alpha}(T) =\frac{V_{\mathrm{MEO}} }{R_{\mathrm{MEO}} }$ with $V_{\mathrm{MEO}} = \sqrt{\frac{GM_{\mathrm{Earth}}}{R_{\mathrm{MEO}}}} = 6.90~\mathrm{km/s}$, and the initial and terminal radial velocity was constrained to zero, $\dot{r}(0)= \dot{r}(T) = 0$. Lastly, the TOF is treated as optimizable and was initialized at 
$T_{\mathrm{init}} = \pi \sqrt{\frac{(R_{\mathrm{LEO}}+R_{\mathrm{MEO}})^{3}}{8GM_{\mathrm{Earth}}}} \approx 3315~\mathrm{s}$.

\begin{figure}[htbp]
\begin{subfigure}[htbp]{.49\linewidth}
    \centering
    \includegraphics[width=1\linewidth]{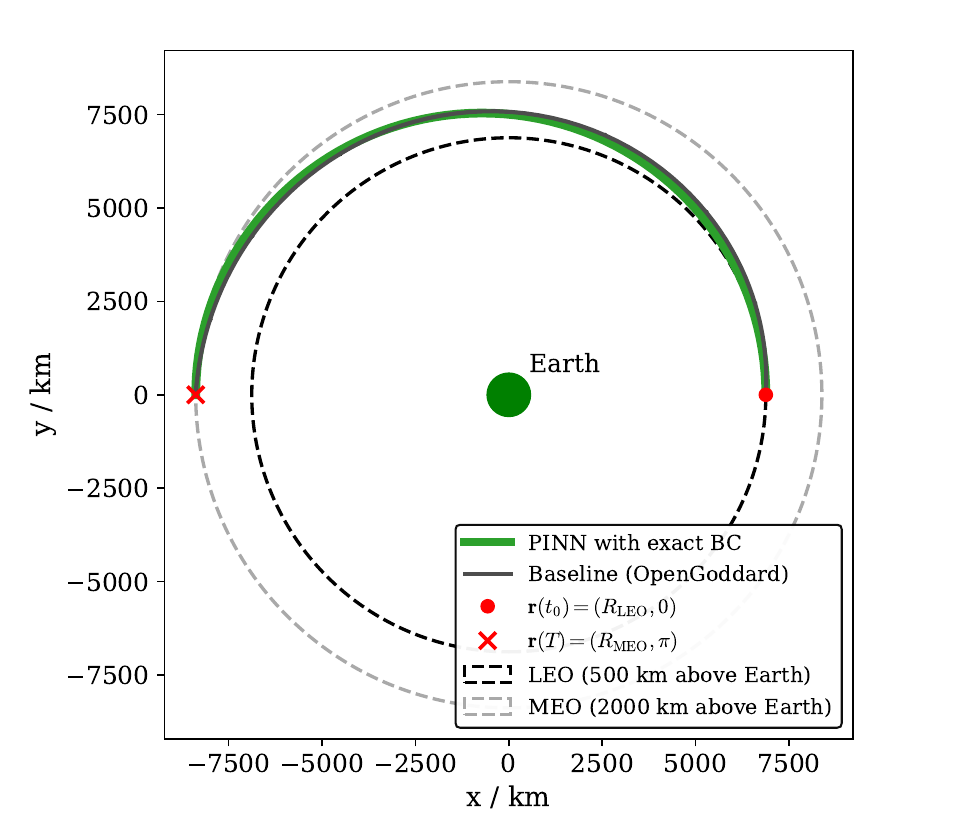}
    \caption{Trajectories.}
    \label{fig:kinematic_ht_traj3d}
    \end{subfigure}
    \begin{subfigure}[htbp]{.49\linewidth}
    \centering
    \includegraphics[width=1\linewidth]{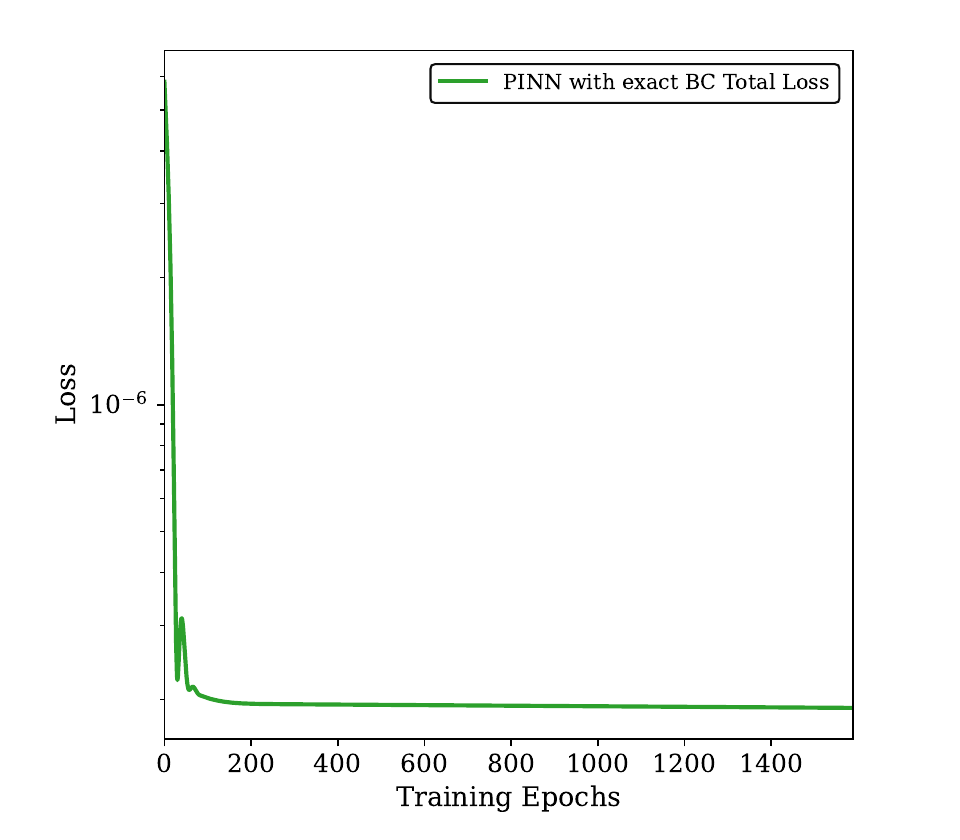}
    \caption{Training curve.}
    \label{fig:kinematic_ht_loss}
    \end{subfigure}

    \begin{subfigure}[htbp]{.49\linewidth}
    \centering
    \includegraphics[width=1\linewidth]{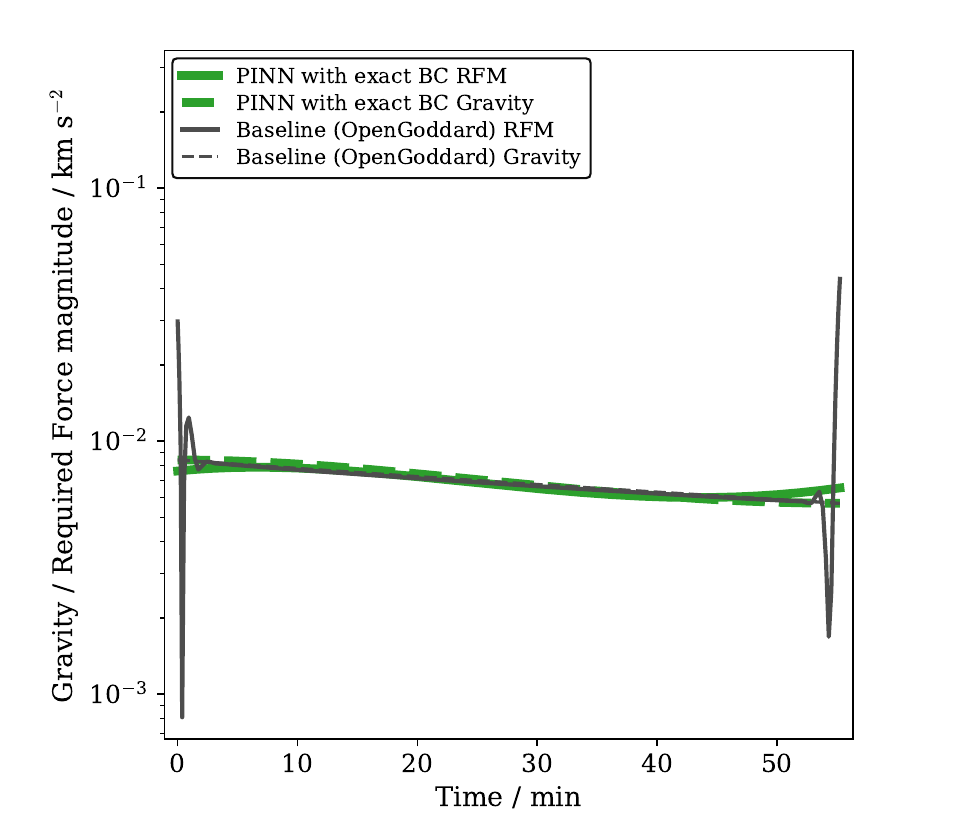}
    \caption{Gravity and required force magnitude (RFM).}
    \label{fig:kinematic_ht_gravity}
    \end{subfigure}
    \begin{subfigure}[htbp]{.49\linewidth}
    \centering
    \includegraphics[width=1\linewidth]{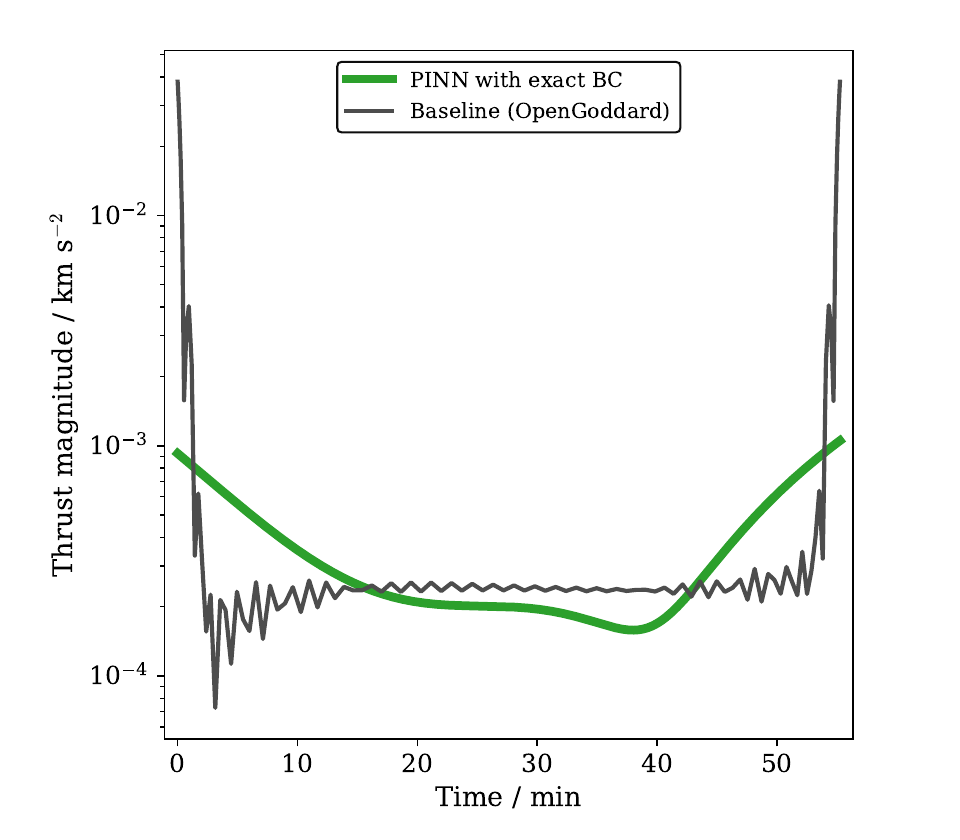}
    \caption{Thrust magnitude.}
    \label{fig:kinematic_ht_thrust}
    \end{subfigure}
    
    \centering
\caption{Circular orbit transfer with fixed terminal angle. 
Comparison of our transformed PINN with exact boundary conditions to the OpenGoddard baseline. }
\label{fig:kinematic_ht}
\end{figure}

The model was trained with ADAM  until a relative loss change of 1e-5.
The results are shown in Fig. \ref{fig:kinematic_ht}. 
For reference, we compare the obtained trajectory to the classical Hohmann transfer, which provides the optimal impulsive $\Delta V$ for transfers between circular coplanar orbits. While the Hohmann solution assumes instantaneous velocity changes, the proposed method yields a continuous low-thrust trajectory. 
The trajectory (top left panel) of the transformed PINN with exact boundary conditions is qualitatively again similar to OpenGoddard's numerical solver baseline. However, in this example the transformed PINN achieves a lower $\Delta V$ than the classical solver, which yields relatively large impulses close to start and end of the trajectory (bottom panels). 
We observe that the transformed PINN solution produces a smooth thrust profile that approximates the impulsive solution while maintaining comparable total control effort. 

We observed that this holds even when we run the classical solver again with a warm re-start where the initialization is based on the transformed PINN's resulting superior trajectory.
The training curve (top right) shows relatively fast convergence of the transformed PINN with exact boundary conditions. Note the naive PINN without exact boundary conditions (i.e. soft BCs) failed to converge to a satisfactory solution and is hence omitted here.

\subsection{Circular Orbit   Transfer with Free Terminal Angle} \label{sec:example4}

We extend the previous example by relaxing the constraints on the phase angle, in particular with free terminal phase, and we consider a transfer from Low Earth Orbit (LEO) to Geostationary Earth Orbit (GEO).
With an initial phase of $\alpha(t_0)=0$, the initial boundary conditions are given by 
$(r(t_0),\alpha(t_0))^T = (R_{\mathrm{LEO}},0)^T$ and $(\dot{r}(t_0),\dot{\alpha}(t_0))^T = (0,V_{\mathrm{LEO}}/R_{\mathrm{LEO}})^T$. At the terminal point, the phase angle is free. Thus, only the orbital radius and tangential velocity are prescribed, i.e. $r(T)=R_{\mathrm{GEO}}$ and $(\dot{r}(T),\dot{\alpha}(T))^T=(0,V_{\mathrm{GEO}}/R_{\mathrm{GEO}})^T$while $\alpha(T)$ is optimized.

The optimization was initialized with a terminal-angle guess of $2.10\,\pi$ and an initial guess for the TOF that equals the Hohmann transfer time multiplied by a factor of $1.09$. This problem proved to be sensitive to the initial guess of the terminal angle as well as the TOF. Hence, a grid search over both quantities was carried out in order to identify a robust initialization region with favorable convergence behavior and low control cost. The model was trained with ADAM  until a relative loss change of 1e-5.

\begin{figure}[htbp]
\begin{subfigure}[htbp]{.49\linewidth}
    \centering
    \includegraphics[width=1\linewidth]{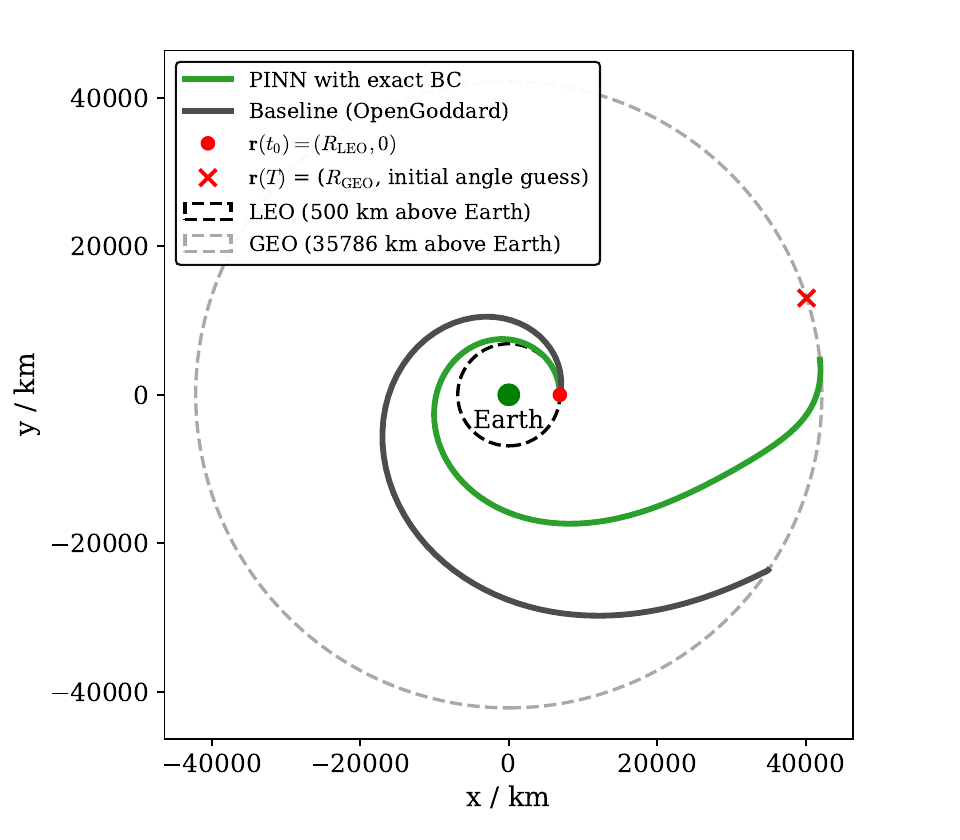}
    \caption{Trajectories.}
    \end{subfigure}
    \begin{subfigure}[htbp]{.49\linewidth}
    \centering
    \includegraphics[width=1\linewidth]{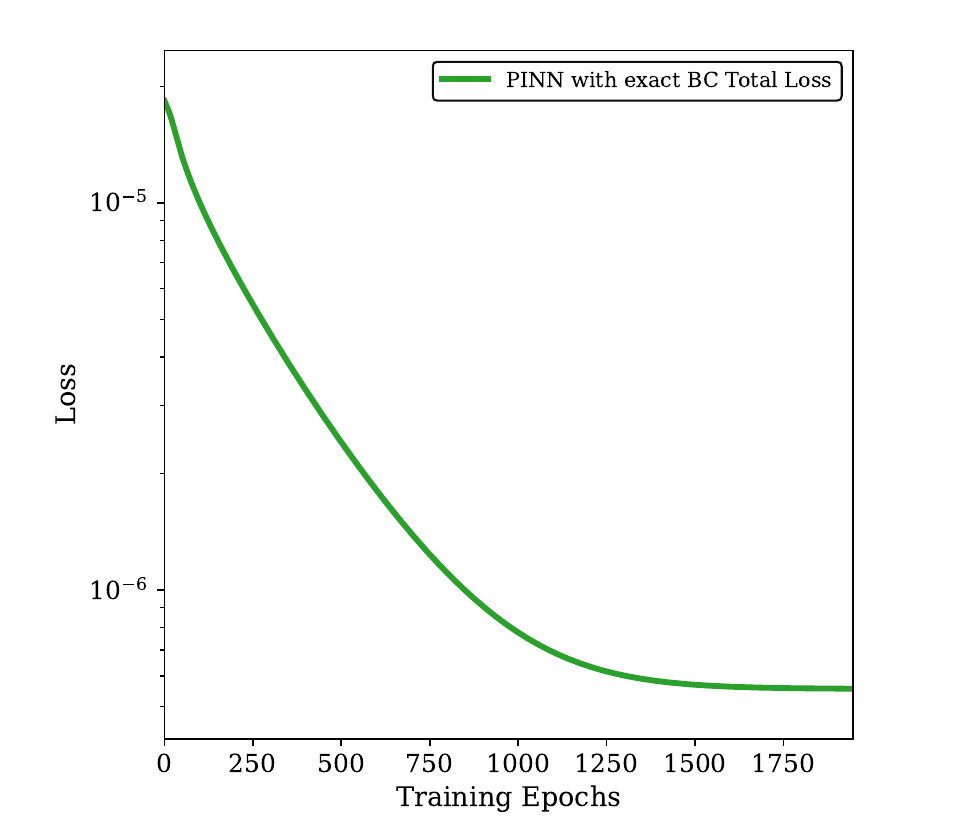}
    \caption{Training curve.}
    \end{subfigure}
    \begin{subfigure}[htbp]{.49\linewidth}
    \centering
    \includegraphics[width=1\linewidth]{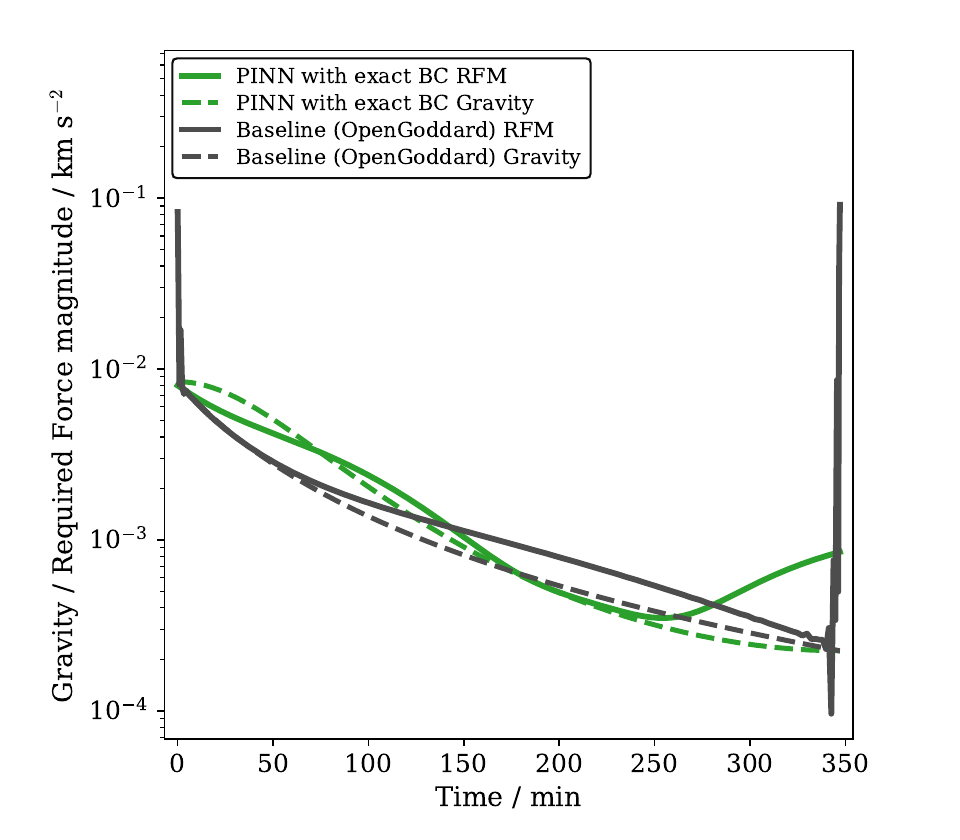}
    \caption{Gravity and required force magnitude (RFM).}
    \end{subfigure}
    \begin{subfigure}[htbp]{.49\linewidth}
    \centering
    \includegraphics[width=1\linewidth]{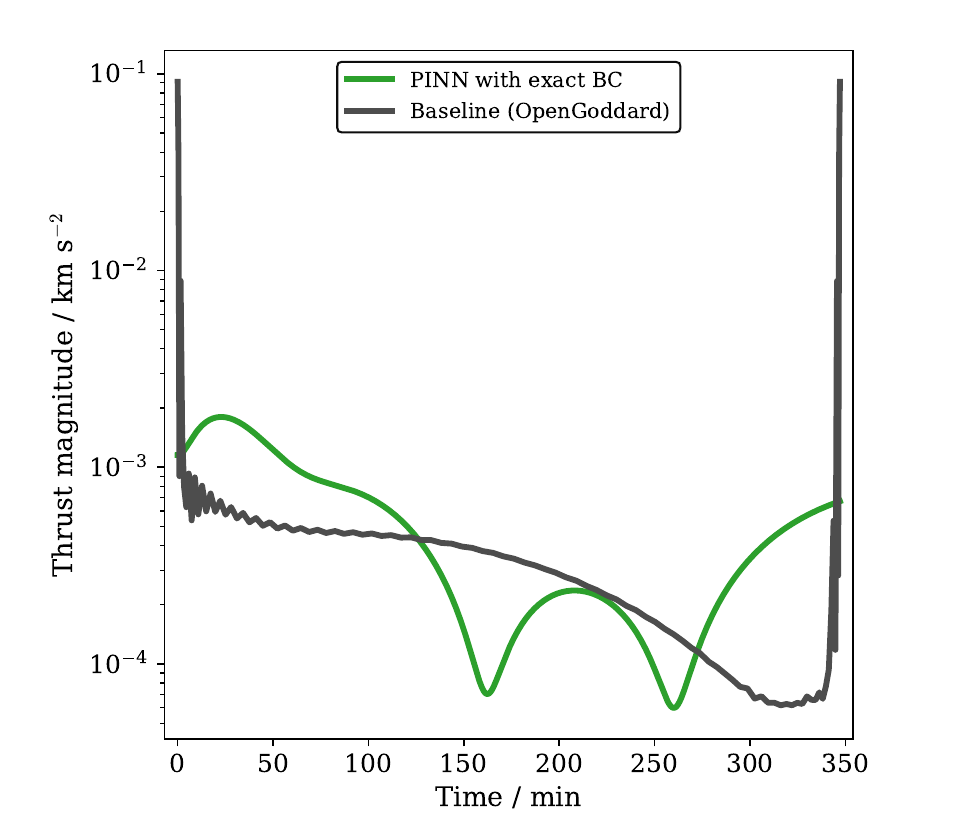}
    \caption{ Thrust magnitude.}
    \end{subfigure}
\caption{Circular orbit transfer with free terminal angle. Comparison of our transformed PINN with exact boundary conditions to the OpenGoddard baseline. }
\label{fig:low_thrust_transfer}
\end{figure}

The results are shown in Fig.~\ref{fig:low_thrust_transfer}. The transformed PINN with exact boundary conditions converged reliably to a smooth low-thrust transfer and produced a trajectory qualitatively similar to the OpenGoddard reference solution. As in the previous example, the naive PINN without exact boundary conditions (i.e. soft BCs) failed to converge to a satisfactory solution. Both trajectories become tangential to the orbit as they close in on the orbit\footnote{This is not obviously visible for the baseline on this figure's scale due to the impulse-like correction happening only shortly before entering orbit.  We omit an additional plot on smaller axes scales for brevity.
}.

\subsection{Hold-Point Rendezvous in Low Earth Orbit} \label{sec:example5}

To complement the orbital-transfer examples by a local proximity-operations scenario, we consider a rendezvous hold-point problem on a circular orbit in the earth-centered inertial frame.
The target follows the same circular LEO reference
orbit as in the previous orbit-transfer example~\ref{sec:example3}.
Introducing subscripts $a$ and $b$ for chaser and target, respectively, in polar
coordinates, its initial state is
$(r_b(t_0),\alpha_b(t_0))^T = (R_{\mathrm{LEO}},0)^T$ and $
(\dot r_b(t_0),\dot\alpha_b(t_0))^T =
\left(0,V_{\mathrm{LEO}}/{R_{\mathrm{LEO}}}\right)^T .$
Although the PINN is implemented in polar coordinates, the rendezvous scenario
is most naturally specified in the target-centered Cartesian frame. Let
$
\mathbf{x}_b(t)=R_{\mathrm{LEO}}~\mathbf e_r(t)
$ and $
\dot{\mathbf{x}}_b(t)=V_{\mathrm{LEO}}~\mathbf e_\alpha(t)
$
denote the target position and velocity, where \(\mathbf e_r(t)\) and
\(\mathbf e_\alpha(t)\) are the radial and along-track unit vectors of the
target orbit. The chaser position is represented relative to the target as
$
\mathbf{x}_a(t)
=
\mathbf{x}_b(t)
+
\delta r_a(t)\mathbf e_r(t)
+
\delta s_a(t)\mathbf e_\alpha(t)
$ where $\delta r_a(t)$ and $\delta s_a(t)$ are the radial and along-track relative offsets, respectively.
The initial relative offset is chosen as
$\delta r_a(t_0) = 
-1.0\,\mathrm{km}$ and $\delta s_a(t_0) = 
-0.3\,\mathrm{km}$ with $
\dot{\mathbf{x}}_a(t_0)=\dot{\mathbf{x}}_b(t_0).
$
The terminal point is defined as a hold point with a prescribed radial offset $h=30~$m outside the target orbit 
$
\mathbf{x}_a(T)
=
\mathbf{x}_b(T)+h\mathbf e_r(T)
$. This corresponds to matching the target's angular rate at the terminal point $
\dot{\mathbf{x}}_a(T)
=
(R_{\mathrm{LEO}}+h)\frac{V_{\mathrm{LEO}}}{R_{\mathrm{LEO}}}
\mathbf e_\alpha(T).
$
For training, these Cartesian boundary states are
converted to polar coordinates \((r,\alpha)\) and  polar velocities
\((\dot{r},\dot{\alpha})\).
Altogether, the problem is solved as a free-TOF low-thrust rendezvous-to-hold-point transfer with exact boundary conditions. The model is trained with ADAM until a relative loss change of 10e-7.

\begin{figure}[htbp]
\begin{subfigure}[htbp]{.49\linewidth}
    \centering
    \includegraphics[width=1\linewidth]{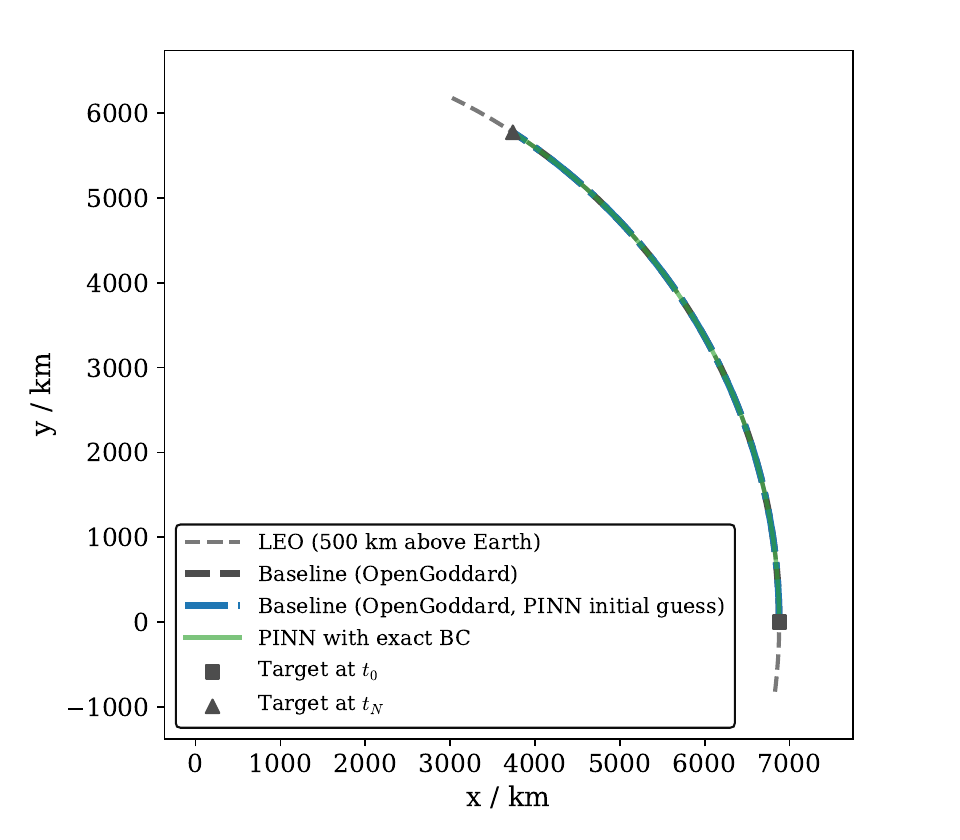}
    \caption{Orbit frame overview and trajectories.}
    \end{subfigure}
    \begin{subfigure}[htbp]{.49\linewidth}
    \centering
    \includegraphics[width=1\linewidth]{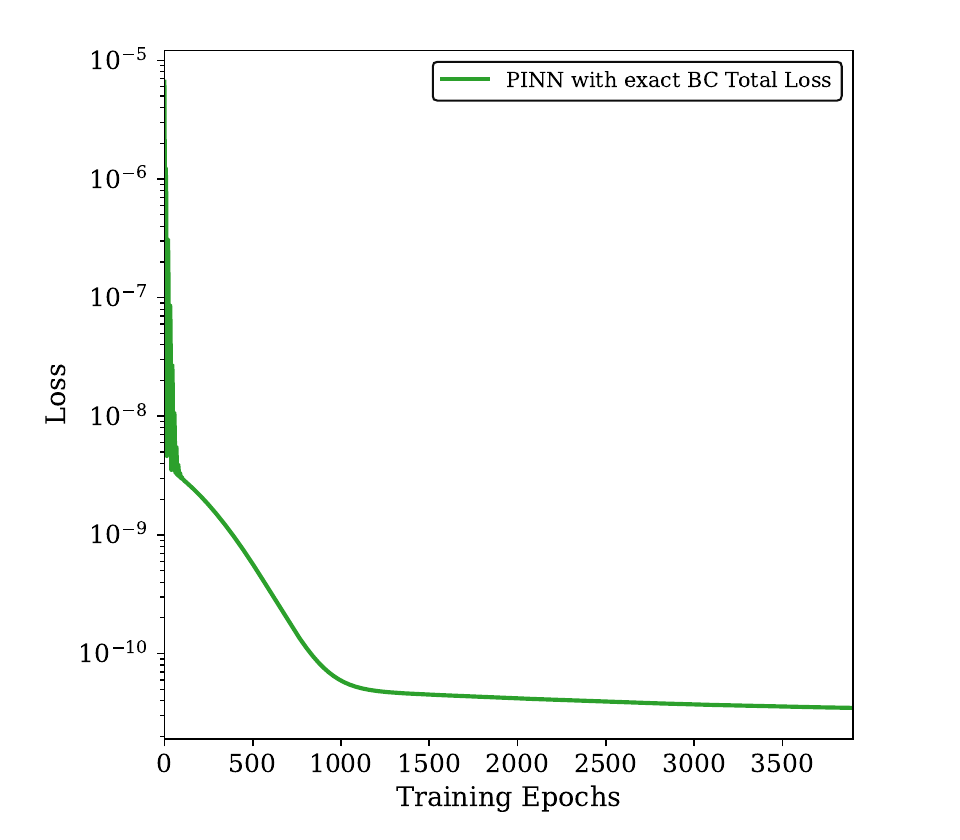}
    \caption{Training curve.}
    \end{subfigure}

    \begin{subfigure}[htbp]{.49\linewidth}
    \centering
    \includegraphics[width=1\linewidth]{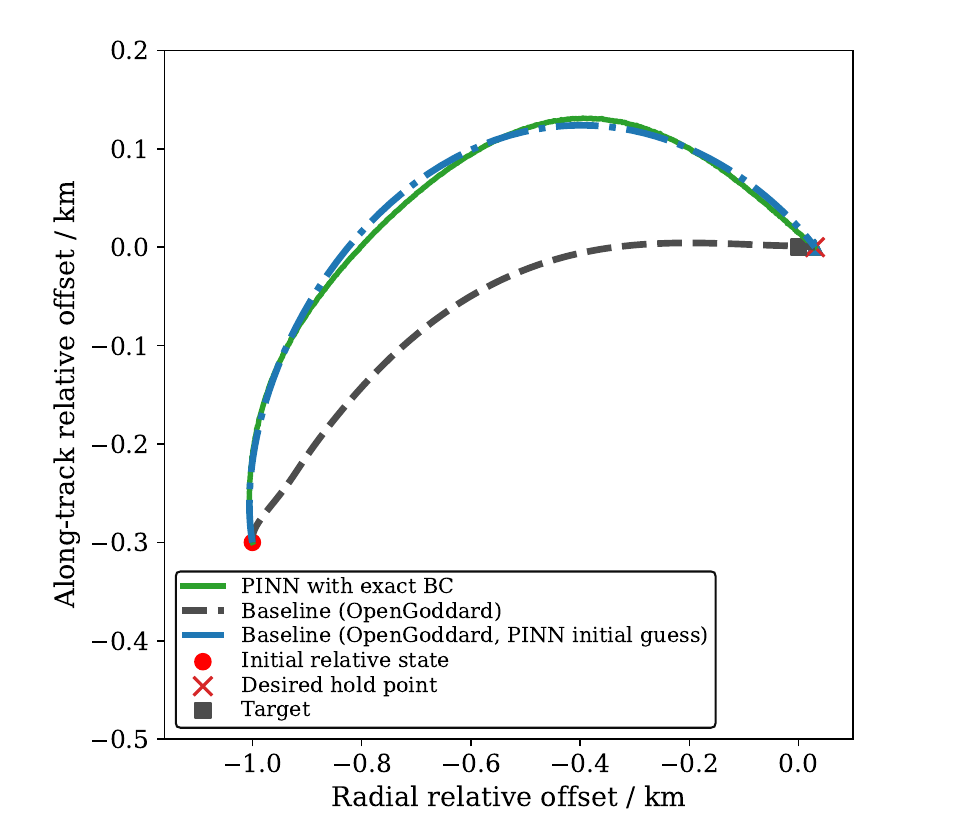}
    \caption{LVLH Relative Motion, target centered at the origin.}
    \end{subfigure}
    \begin{subfigure}[htbp]{.49\linewidth}
    \centering
    \includegraphics[width=1\linewidth]{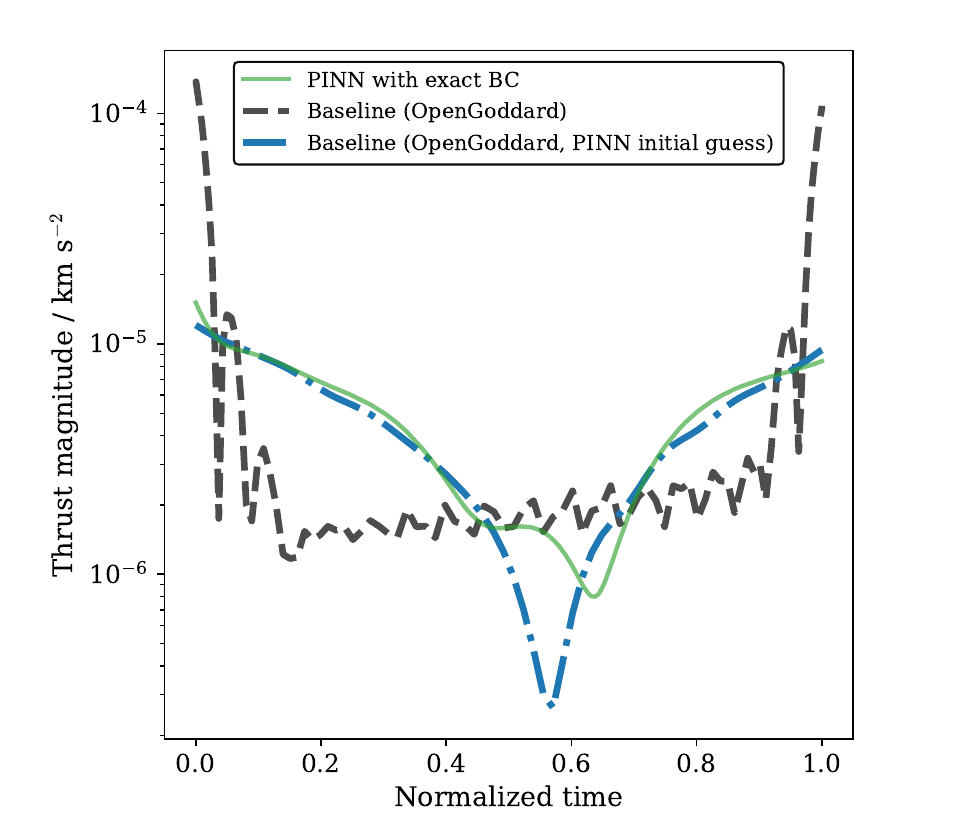}
    \caption{Thrust magnitude.}
    \end{subfigure}
\caption{Hold-Point Rendezvous in LEO with initial offset between target and chase.}
\label{fig:rendezvous}
\end{figure}

The results are shown in Fig. \ref{fig:rendezvous}, including an orbit-frame overview of the global motion, a local vertical local horizontal (LVLH) relative-motion plot, a thrust-magnitude history in minutes and the total loss convergence plot.  The transformed PINN successfully computes a low-thrust rendezvous trajectory to the prescribed hold point, satisfying all boundary conditions exactly and exhibiting stable convergence without the need for boundary-condition loss tuning.

\subsection{Summary metrics and statistical analyses} \label{sec:summary}
\begin{figure}[htbp]
    \centering

    \begin{subfigure}{1\textwidth}
        \centering
        \includegraphics[width=\linewidth]{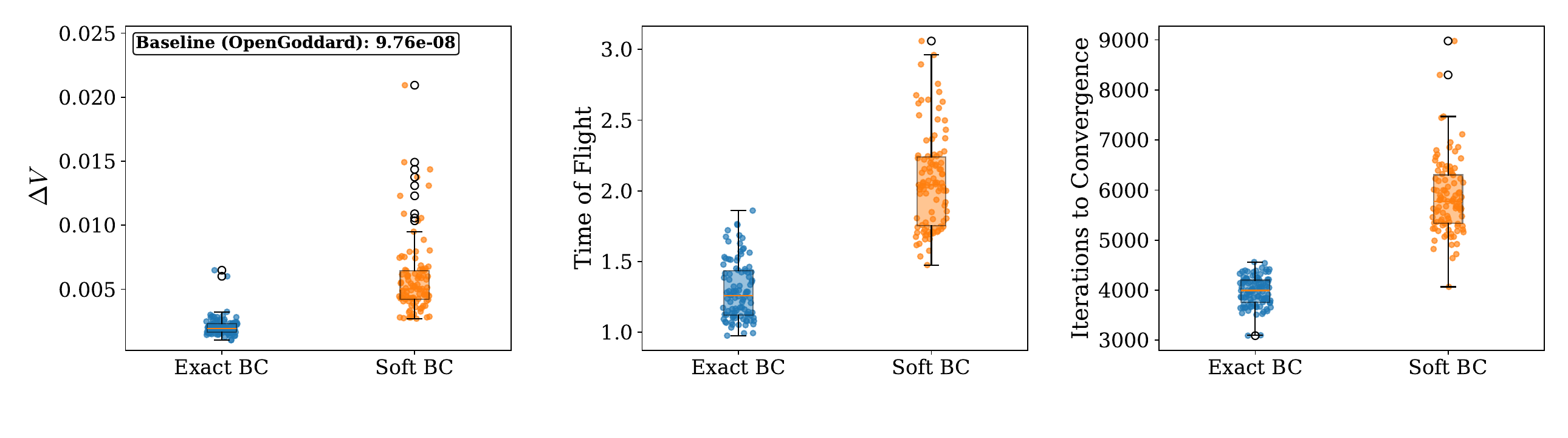}
        \caption{
        Swing-by in 2D (Sec. \ref{sec:example1}). From left to right: $\Delta V$, TOF, training iterations until convergence.}
        \label{fig:sub1}
    \end{subfigure}

    \vspace{0.5cm}

    \begin{subfigure}{1\textwidth}
        \centering
        \includegraphics[width=\linewidth]{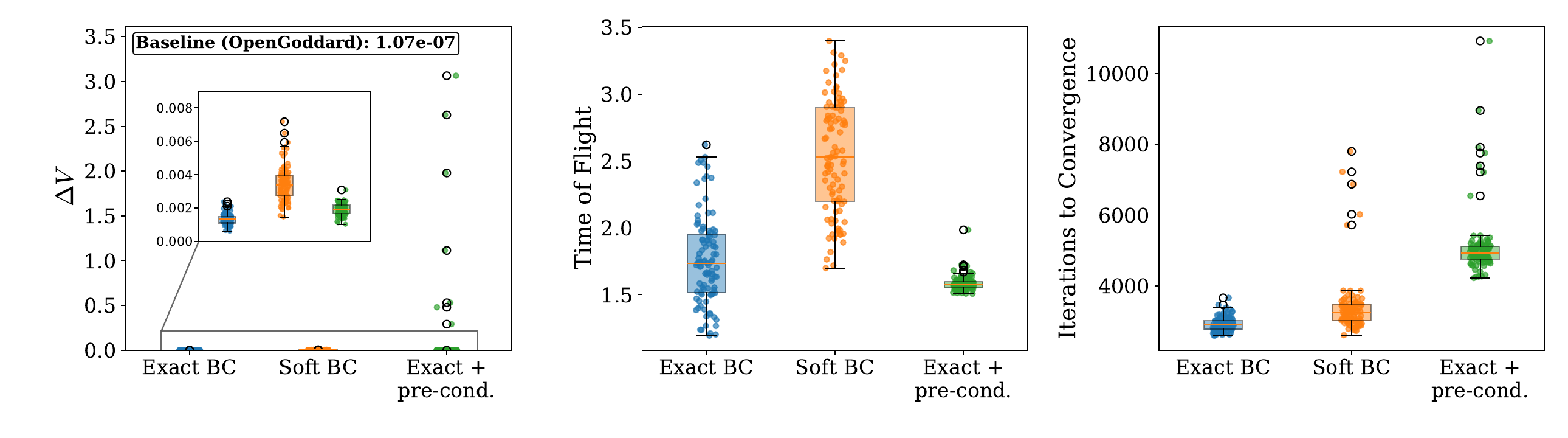}
        \caption{Swing-by in 3D (Sec. \ref{sec:example2}). From left to right: $\Delta V$, TOF, training iterations until convergence. }
        \label{fig:sub2}
    \end{subfigure}

    \caption{
    Box plots for 100 randomly initialized PINNs, comparing naive PINNs with soft BCs to our transformed PINNs with exact BCs, and a pre-conditioned variant of the latter.
    }
    \label{fig:main-box-plot}
\end{figure}

The $\Delta V$ of all above examples is summarized for all methods in Tab. \ref{tab:comparison_methods}. Furthermore, we show box plots of $\Delta V$ for 100 randomly initialized neural networks in Fig. \ref{fig:main-box-plot}, to demonstrate robustness and consistency of performance. We limit the statistical comparison to the examples in Sec. \ref{sec:example1} and Sec. \ref{sec:example2} since for the advanced examples (Sec. \ref{sec:example3}, Sec. \ref{sec:example4} and Sec. \ref{sec:example5}), the naive PINN with soft BCs failed to converge.

In 2D, the transformed PINN with exact BC consistently outperforms PINN with soft BC. In 3D, the transformed PINN with exact BC again improves over PINN with soft BC, while the pre-trained variant attains competitive median performance but exhibits a small number of outliers that failed to converge. The exact-BC formulation achieves lower $\Delta V$, shorter TOF, and faster convergence across most random seed initializations. The pre-trained variant improves median performance but introduces heavy tails, visible in the $\Delta V$ and convergence distributions. We find consistently that the naive PINN underperforms or fails because it must trade off objectives, whilst the transformed PINN removes that trade-off entirely. By restricting the function space to feasible trajectories, the transformation modifies the optimization landscape and effectively removes infeasible local minima that arise from boundary condition violations. Mind that different methods can converge to other distinct local optima -- not related to the boundary conditions, nevertheless -- resulting in additional outliers.  

\begin{table}[t]
\centering
\caption{Summary metrics and comparisons for the five numerical examples Sec. \ref{sec:example1} to Sec. \ref{sec:example5}. We used 100 samples each. PINNs with soft BCs did not converge in some examples (denoted as n.c.).}
\begin{tabular}{rlrrr}
\hline\hline
Example & Method & mean of  $\Delta V$ & variance of $\Delta V$ & CPU-time / s \\
\hline
Swing-by 2D   & OpenGoddard (baseline) & $9.76\times10^{-8}$ & - & $34.72$\\
(Sec. \ref{sec:example1})    & PINN with soft BCs$^*$          & $5.84\times10^{-3}$ & $8.76\times10^{-6}$ & $30.70$\\
    & PINN with exact BCs & $2.06\times10^{-3}$ & $5.60\times10^{-7}$ & $23.51$\\ \hline
Swing-by 3D   & OpenGoddard (baseline) & $1.07\times10^{-7}$ & - & $113.57$\\
Sec. \ref{sec:example2}    & PINN with soft BCs          & $3.43\times10^{-3}$ & $1.11\times10^{-6}$ & $24.04$\\
    & PINN with exact BCs & $1.35\times10^{-3}$ & $1.06\times10^{-7}$ & $26.26$\\
    & PINN with exact BCs + pre-conditioning$^{**}$ & $1.88\times10^{-3}$ & $1.24\times10^{-7}$ & $25.95$\\ \hline

Orbit transfer w/  & OpenGoddard (baseline) & $2.17$ & - & $577.57^{\dagger}$\\
 fixed terminal angle    & Analytical (Hohmann transfer) & $0.71$ & - & -\\
(Sec. \ref{sec:example3})    & PINN with soft BCs         & n.c. & n.c. & n.c. \\
    & PINN with exact BCs & $1.21$ & $2.97\times10^{-5}$ & $2.05$\\    \hline
Orbit transfer w/   & OpenGoddard (baseline) & $12.03$ & - & $413.06^{\dagger}$ \\
 free terminal angle    & PINN with soft BCs          & n.c. & n.c. & n.c.\\
(Sec. \ref{sec:example4})    & PINN with exact BCs & $11.60$ & $2.54\times10^{-3}$ & $9.63$\\\hline
Rendezvous     & OpenGoddard (baseline) & $5.51\times10^{-3}$ & - & $489.50^{\dagger}$\\
 (Sec. \ref{sec:example5})    & PINN with exact BCs & $4.40\times10^{-3}$ & $9.74\times10^{-9}$ & $5.35$\\ 
    & OpenGoddard w/ PINN as initial guess$^{***}$ & $4.25\times10^{-3}$ & - & $487.65^{\dagger}$\\
    
    \hline
\hline
\hline
\end{tabular}
\vspace{0.5em}
\parbox{0.98\linewidth}{\footnotesize $^{*}$ One outlier run was excluded from the reported statistics.} 
\parbox{0.98\linewidth}{\footnotesize $^{**}$ Seven outlier runs were excluded from the reported statistics.}
\parbox{0.98\linewidth}{\footnotesize $^{***}$ Specifically, the solution by PINN with exact BCs was input as initial guess for a new solve with OpenGoddard.}
\parbox{0.98\linewidth}{\footnotesize $^{\dagger}$ Baseline optimization stopped after reaching iteration budget.}

\label{tab:comparison_methods}
\end{table}

\section{Discussion}
\label{sec:discussion}

The numerical results presented in Sec.~\ref{sec:experiments} demonstrate that the proposed transformation-based PINN formulation enables exact enforcement of boundary and kinematic constraints while substantially improving convergence behavior compared to ordinary PINNs. Across all test cases, transformed PINN with exact boundary conditions (BCs) either (i) consistently achieved lower residual losses, faster convergence, and smoother thrust profiles without requiring adjustment of boundary-loss weights, or (ii) enabled converged solutions in the first place. 

The results imply a fundamental limitation of standard PINN formulations for trajectory optimization: the need to balance competing objectives in a multi-objective loss function. The introduction of boundary-condition penalty terms leads to a sensitivity with respect to loss weights, which can significantly affect both convergence and solution quality. By eliminating these penalty terms entirely, the proposed method removes a key source of instability and, in the absence of additional objectives or constraints, reduces the training problem to a single-objective optimization.

From a methodological perspective, the proposed transformation can be interpreted as a reparameterization of the solution space that embeds the boundary conditions directly into the function class. This converts the original constrained optimization problem into an unconstrained optimization over admissible trajectories. This viewpoint explains the improved robustness and reduced hyperparameter sensitivity observed in the experiments and connects the approach to classical ideas in constrained optimization.

In comparison to classical trajectory optimization methods, such as direct collocation, the proposed approach offers a fundamentally different trade-off. While classical methods operate on discretized trajectories and require explicit enforcement of constraints at collocation points, the present formulation yields a continuous representation of the trajectory that is (auto-)differentiable with respect to both time and problem parameters. This enables gradient-based optimization in function space and provides a natural pathway toward solving inverse problems and sensitivity analysis within the same framework.

The comparison with the Hohmann-type transfer illustrates that the proposed method can recover physically meaningful solutions that are consistent with classical astrodynamics. While the Hohmann transfer represents an optimal impulsive solution, the PINN-based approach produces a smooth, continuous low-thrust trajectory that approximates this behavior without requiring explicit discretization of control inputs. This suggests that the PINN solution can be interpreted as a continuous generalization of classical $\Delta V$-optimal transfers.

The observed performance gains were more pronounced for increasingly complex dynamical systems, where the ordinary PINN often failed to converge, converged to the \lq\lq wrong\rq\rq~solution or required extensive tuning of $\lambda_{\mathrm{BC}}$. In contrast, the transformed PINN with exact boundary conditions maintained reliable convergence even in free TOF optimization. The transformed formulation also integrates seamlessly with existing PINN toolchains, preserving differentiability and using standard optimizers such as ADAM and L-BFGS.

Overall, the results indicate that embedding constraints directly into the neural network parameterization is a promising direction   in trajectory optimization. By combining exact constraint satisfaction with continuous representations and automatic differentiation, the proposed framework provides a flexible   alternative to traditional methods. 

\subsection{Limitations}

At the same time, the experiments reveal several limitations. In higher-dimensional settings, such as the 3D swing-by problem, the optimization landscape may contain multiple, physically meaningful local minima. As a result, the method can benefit from suitable initialization or pre-conditioning strategies, but exact boundary-condition enforcement alone does not guarantee convergence to globally optimal trajectories. This behavior is consistent with the non-convex nature of neural network training and is not unique to the proposed approach; similar challenges arise in classical methods such as direct collocation, albeit with more explicit control through initialization.
 
Beyond this, the present study is limited to relatively simple problems with idealized gravitational models. Extending the approach to realistic astrodynamics scenarios will require incorporating perturbations, eclipses, and higher-order dynamics. Moreover, scalability to long-duration missions may benefit from domain decomposition or adaptive collocation strategies, as the current formulation relies on a single monolithic neural network. While the proposed transformation removes the need for boundary-condition tuning, performance still depends on neural-network architecture and optimizer settings, which warrant further systematic investigation. Although convergence behavior is improved empirically, a formal analysis of convergence rates and numerical conditioning remains an open challenge. Finally, the presence of multiple local optima can lead to qualitatively distinct solution families, not all of which may be desirable from a mission-design perspective. While such local minima often correspond to physically meaningful trajectories, they complicate optimization and require careful initialization or guidance. This challenge is not unique to the proposed approach and also arises in classical methods such as direct collocation, where more explicit convergence control can be achieved through tailored initialization strategies. 

Importantly, the proposed approach directly applies to equality-type kinematic constraints; extensions to inequality-, path- or control constraints such as bounded thrust remains an open direction. 
Overall, the proposed transformation introduces a principled way to embed boundary information directly into the neural representation for trajectory optimization with practically relevant advantages.

\subsection{Outlook}
Our PINN-based approach, incl. the exact imposition of boundary conditions, could also be combined with the  popular reinforcement learning approaches \cite{sullivan2023designing,bonasera2023designing,salmaso2023machine,gaudet2020reinforcement,gaudet2020deep} to a \lq\lq physics-informed reinforcement learning" approach. Another promising direction would be neural operators \cite{lu2021learning}, incl. physics-informed variants of neural operators \cite{li2024physics}.

Operator learning  offers a natural next step for extending the present framework beyond trajectory optimization toward learning entire solution maps between function spaces. Instead of parameterizing a single trajectory, operator-based models can learn mappings from boundary conditions, system parameters, or force fields directly to optimal trajectories. This could enable rapid amortized inference across families of problems and reduce the need to retrain models for each new instance. In combination with exact constraint enforcement as proposed in this work, operator learning could provide a scalable pathway toward real-time, physics-consistent trajectory design.

\section{Conclusion}

In this work, we introduced a simple yet effective transformation of physics-informed neural networks (PINNs) for spacecraft trajectory optimization that enforces boundary conditions exactly by construction. By embedding boundary conditions directly into the function class via distance functions, the proposed approach reduces the need for penalty-based enforcement and reduces the problem to a single-objective formulation. This alleviates a central challenge in standard PINNs, namely sensitivity to loss-weight tuning and training instability.

Across trajectory optimization problems of increasing complexity, including multi-body swing-by maneuvers and low-thrust orbital transfers, we showed improved convergence, robustness, and solution quality compared to naive PINN formulations. The transformed PINN consistently achieves exact boundary satisfaction, avoids residual constraint violations, and remains stable even where standard PINNs fail to converge.  
Moreover, the method can yield competitive performance relative to classical solvers while offering additional flexibility through auto-differentiability with respect to physical parameters such as the TOF and without requiring fixed discretizations.

At the same time, several challenges remain. Performance still depends on architectural and optimization choices, and complex problems may exhibit multiple physically meaningful local optima, necessitating suitable initialization or pre-conditioning strategies. Extending the framework to more realistic astrodynamics scenarios with perturbations and operational constraints is an important next step. Finally, a rigorous theoretical analysis of convergence and numerical conditioning remains an open question.

More broadly, the proposed approach can be viewed as a constraint-embedding strategy for scientific machine learning. By reparameterizing the solution space to satisfy constraints a priori, it provides a principled alternative to penalty-based formulations that removes a fundamental failure mode of PINNs.

We anticipate that this perspective can contribute to improving robustness and reliability of physics-informed learning methods in engineering and the physical sciences.

\appendix
\section*{Appendix}
\section{Hyperparameter tuning} \label{app:hyperparameter-tuning}
\begin{figure}[htbp]
    \centering
    \begin{subfigure}[b]{0.49\linewidth}
        \centering
        \includegraphics[width=\linewidth]{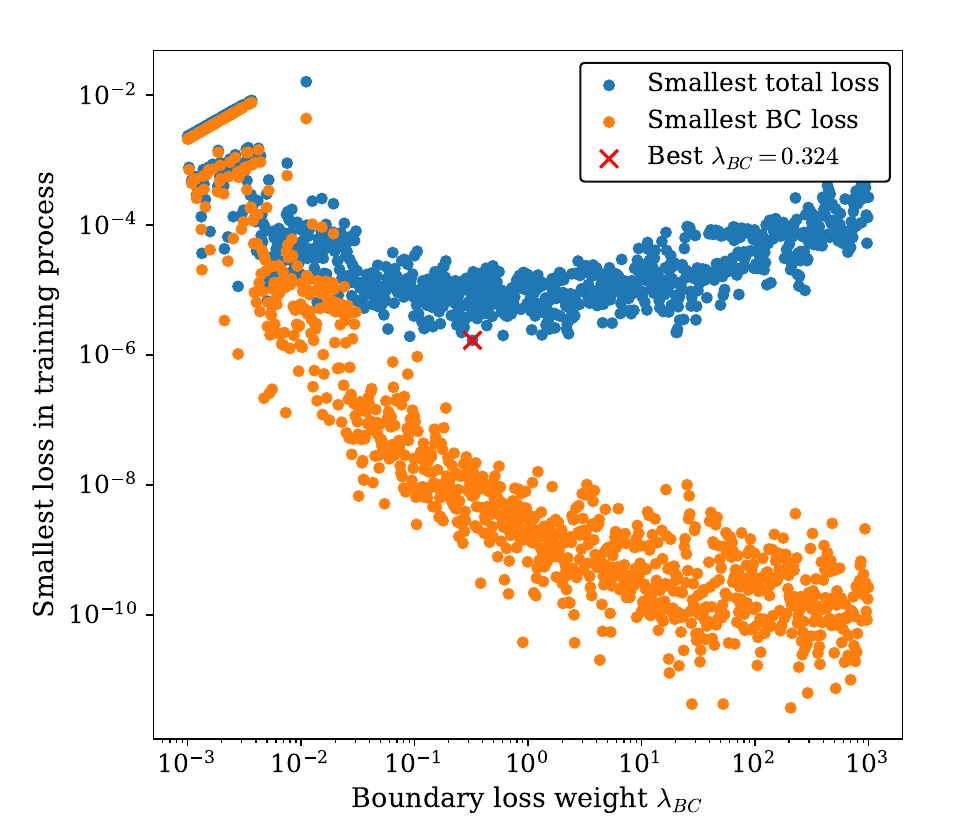}
        \caption{2D problem (Sec. \ref{sec:example1}) with optimizable TOF}
        \label{fig:weight2d_dynamic}
    \end{subfigure}
    \begin{subfigure}[b]{0.49\linewidth}
        \centering
        \includegraphics[width=\linewidth]{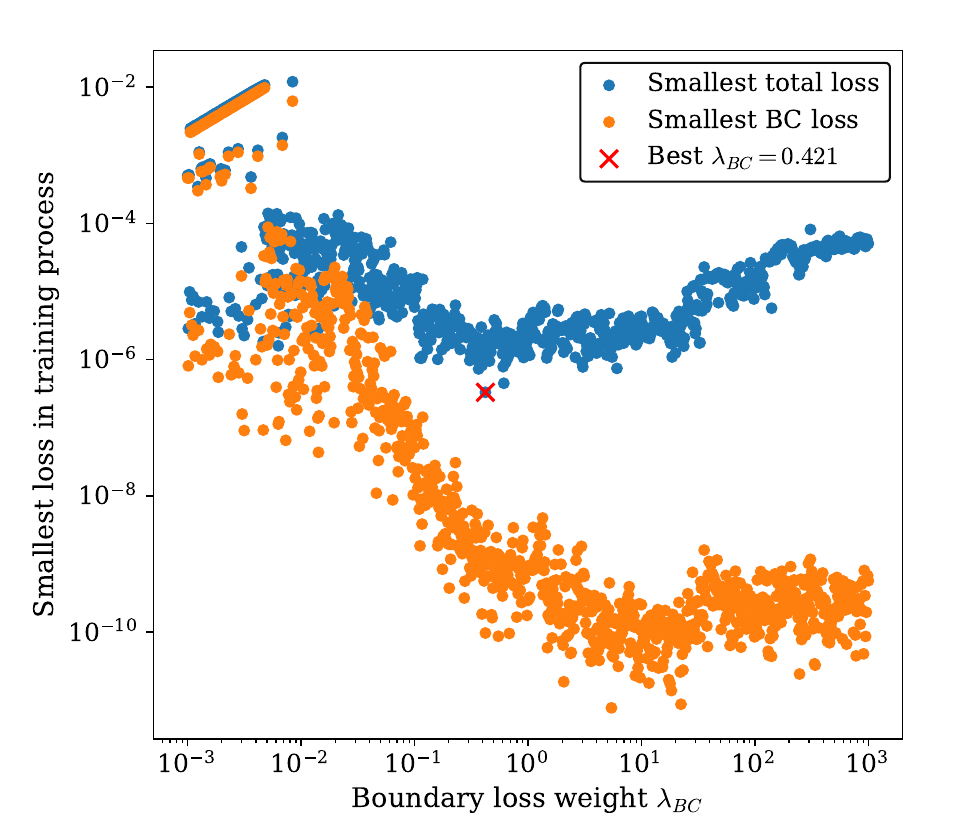}
        \caption{3D problem (Sec. \ref{sec:example2}) with optimizable TOF}
        \label{fig:weight3d_dynamic}
    \end{subfigure}
    \caption{Boundary loss weight $\lambda_{BC}$ grid search for the naive PINNs. Best total loss values are marked with crosses.}
    \label{fig:hyperparameter}
\end{figure}

For fair comparison between an ordinary PINN and a transformed PINN with exact boundary conditions, the only tunable hyperparameter is the loss-weight $\lambda_{BC}$ pertaining to the boundary condition-term in the loss function and setting $\lambda_{BC} = \lambda_i, ~~ \forall i \in \{1,2,3,4\}$. This was achieved via a simple grid search; training 1000 models in the range of $[10^{-3},10^3]$. The results are shown in Fig. \ref{fig:hyperparameter} for both swing-by examples in 2D and 3D, Sec. \ref{sec:example1} and \ref{sec:example2}. 

The PyTorch LBFGS optimizer was used with parameter \texttt{max\_iter=10}, corresponding to at most 10 internal quasi-Newton iterations per optimizer step. The settings for the optimizers and resulting hyperparameters are shown in Table \ref{tab:optimization-settings}.  We omit the corresponding results for the orbital transfer examples (Sec. \ref{sec:example3} and \ref{sec:example4}) due to failure to find a solution that converges to a satisfying solution at all.

The resulting hyperparameter search is visualized in Fig. \ref{fig:hyperparameter}. The blue curve shows the smallest total loss value of a training process with $N_{ADAM}$ = 2000 and at maximum $N_{LBFGS}$ = $10^5$ iterations, for a given $\lambda_{BC}$ and trainable TOF, initialized with $T=1$. The networks were trained 1000 times, respectively. 
We observe in Fig. \ref{fig:hyperparameter} that the ordinary PINN without exact boundary conditions (i.e. soft BCs) is sensitive to the choice of $\lambda_{BC}$. We also observe that the minimum of the total loss does not necessarily need to coincide with the minimum of the boundary condition-loss term alone.

\begin{table}
    \centering
    \caption{Optimizer parameters and results of boundary loss-weight ($\lambda_{BC}$) grid search. From left to right: problem dimension, total TOF $T$ set to fixed or optimizable, number of training iterations with ADAM, maximum number of training iterations with LBFGS optimizer, respectively, and the best resulting boundary weight $\lambda_{BC}$. }
    \begin{tabular}{lllll}

    \hline\hline
      Dim. & $T$ & $N_{ADAM}$ & $\max(N_{LBFGS})$ & Best $\lambda_{BC}$ \\ \hline
        2 & Optimizable & 2000 & 100000 & 0.324 \\ 
        3 & Optimizable & 2000 & 100000 & 0.421 \\ \hline\hline
        \label{tab:optimization-settings}
    \end{tabular}
    \label{tab:grid_search}
\end{table}

\section{Influence of TOF} \label{app:C}
Before, the experiment was conducted with a free terminal time or time-of-flight (TOF) $T$ according to \eqref{eq:minimization-problem-def-trajectory-terminal-time}, where the initial guess for $T$ at the start of the optimization was $T=1$.  
Next we want to investigate the influence of the TOF parameter on the optimization and solution. For this, we consider a variant of the experiment in Sec. \ref{sec:example2} with fixed TOF $T$ and conduct a parameter sweep over $T$ in the range $T\in\left[1,1.1\right]$ in increments of $0.02$ units as well as for discrete values $T\in \{1.0,1.25,1.5,1.75,2.0,3.0\}$. The purpose of this particular set of parameter values is to better visualize a bifurcation with respect to $T$. 

\begin{figure}[htbp]
    \centering
    \begin{subfigure}[b]{0.45\textwidth}
        \includegraphics[width=\textwidth]{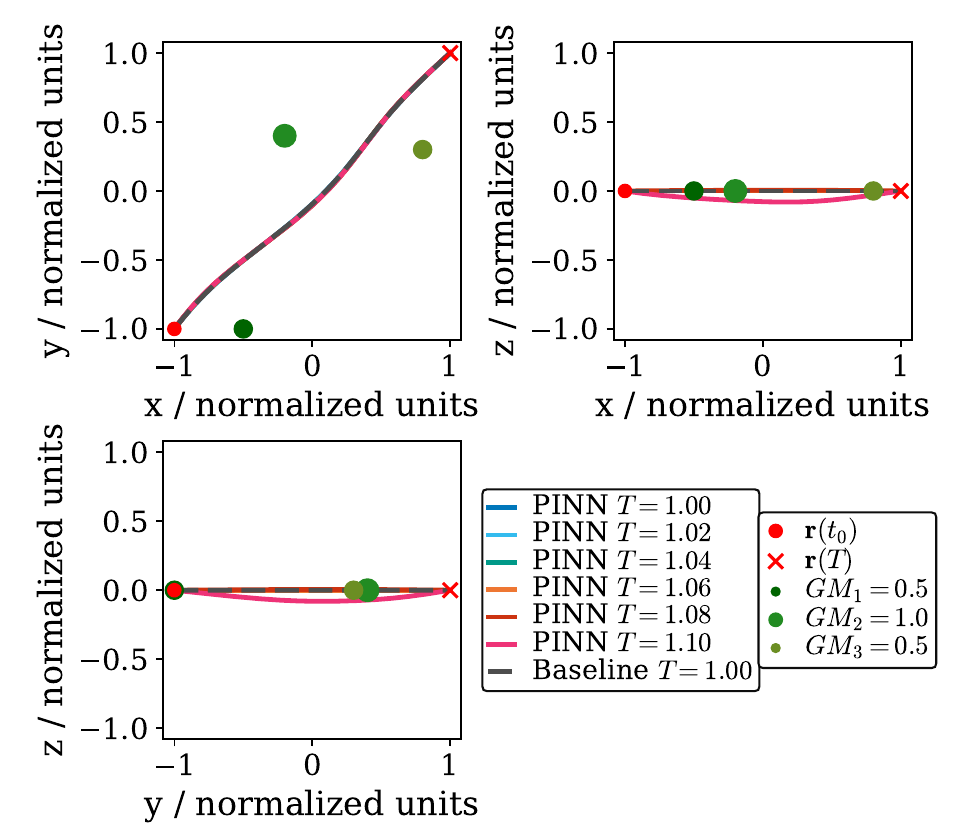}
        \caption{}
        \label{fig:3d_traj_flat}
    \end{subfigure}
    \begin{subfigure}[b]{0.45\textwidth}
        \includegraphics[width=\textwidth]{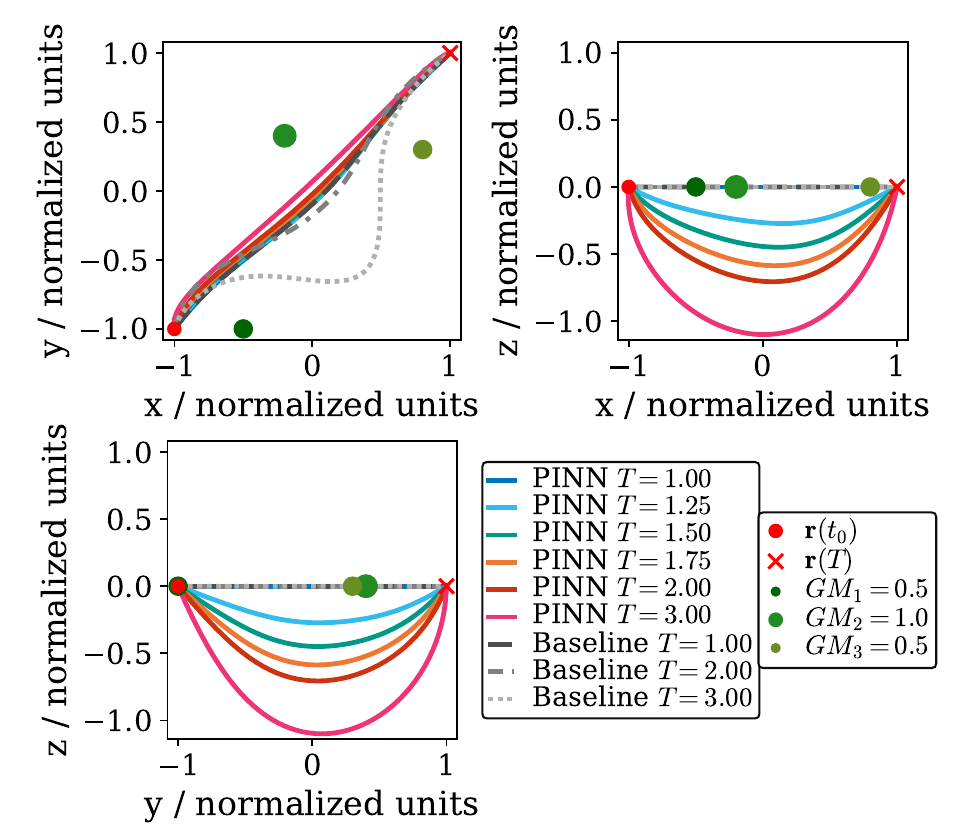}
        \caption{}
        \label{fig:3d_traj_escape}
    \end{subfigure}

    \begin{subfigure}[b]{0.45\textwidth}
        \includegraphics[width=\textwidth]{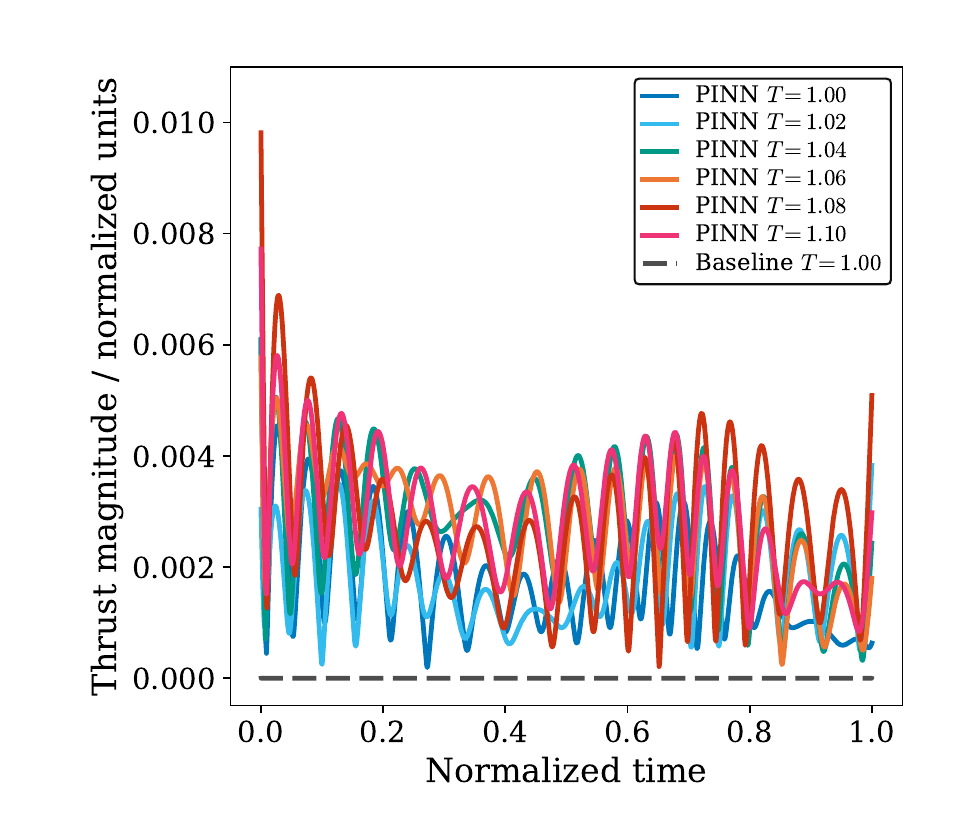}
        \caption{}
        \label{fig:thrust_flat}
    \end{subfigure}
    \begin{subfigure}[b]{0.45\textwidth}
        \includegraphics[width=\textwidth]{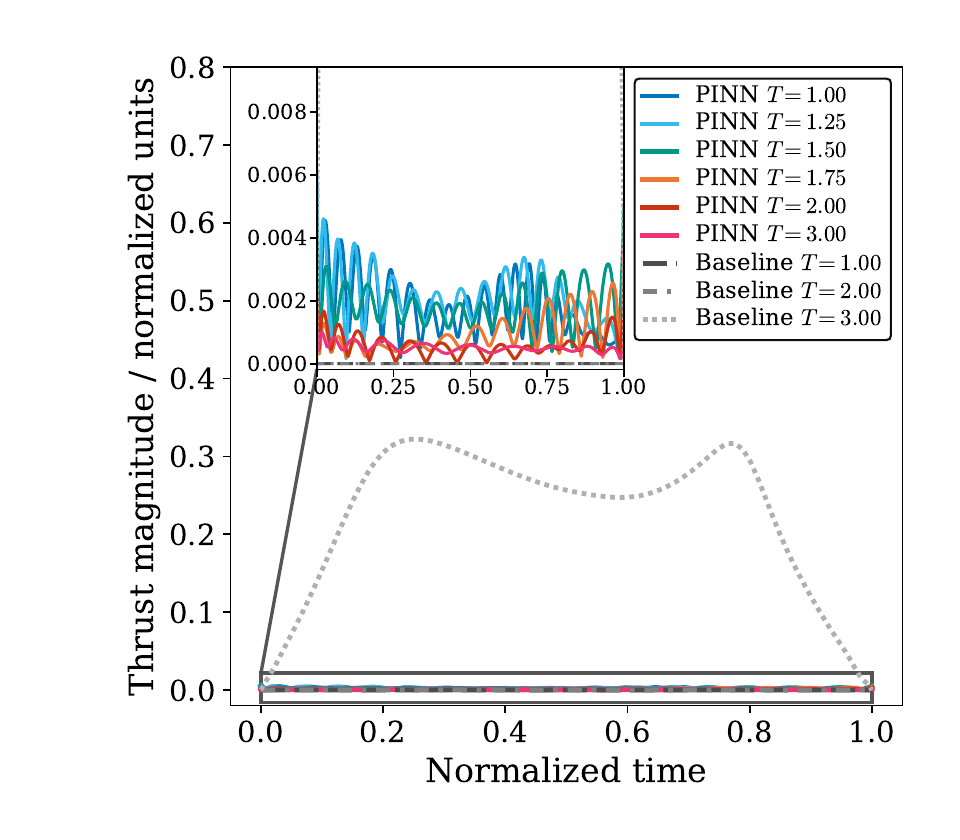}
        \caption{}
        \label{fig:thrust_escape}
    \end{subfigure}

    \begin{subfigure}[b]{0.45\textwidth}
        \includegraphics[width=\textwidth]{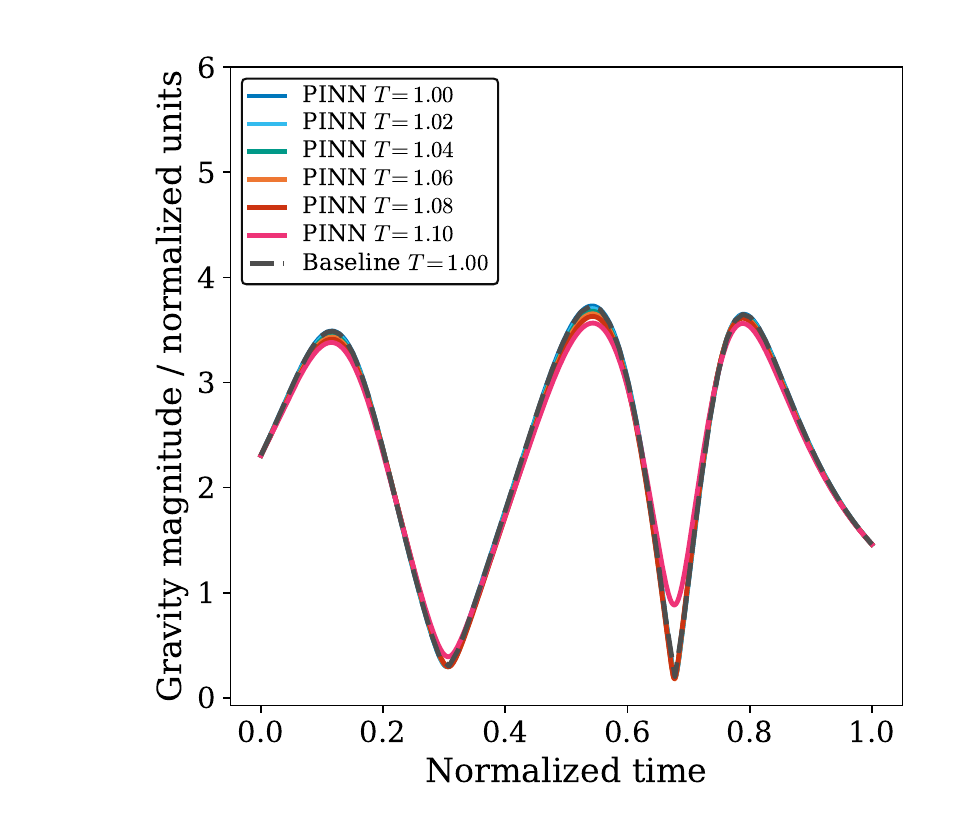}
        \caption{}
        \label{fig:gravity_flat}
    \end{subfigure}
    \begin{subfigure}[b]{0.45\textwidth}
        \includegraphics[width=\textwidth]{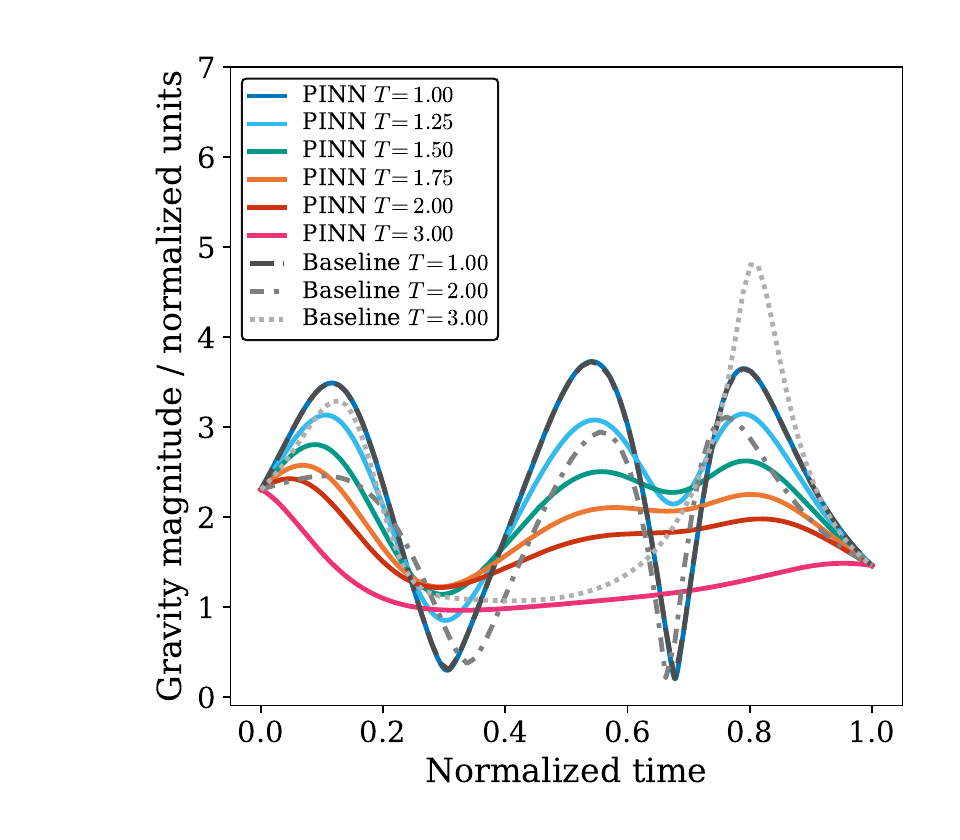}
        \caption{}
        \label{fig:gravity_escape}
    \end{subfigure}

    \caption{Influence of TOF: Parameter sweep for fixed TOF values. Left column: $T\in \left[1.0,1.1\right]$ in increments of 0.02, resulting in direct trajectories on or close to the $(x,y)$-plane. Right column: $T\in \{1.0, 1.25, 1.5, 1.75, 2.0, 3.0\}$ resulting in detour trajectories leaving the $(x,y)$-plane. Top row: Resulting trajectories. Middle row: resulting thrust magnitude. Bottom row: Gravitational pull experienced by the spacecraft.}
    \label{fig:TOF_sweep}
\end{figure}

The results are shown in Fig. \ref{fig:TOF_sweep}. We observe from Fig. \ref{fig:3d_traj_flat} and \ref{fig:3d_traj_escape} that a bifurcation is present roughly in the vicinity of $T\in[1.08,1.1]$. As expected, for larger TOFs, the detour out-of-plane trajectory encountered before is favorable over a direct trajectory, and the arc length of the detour trajectory increases with fixed TOF. This confirms physical consistency in that constraining the TOF to smaller values necessitates a shorter arc length whilst unconstrained TOF translates to virtually no restrictions on the trajectory's arc length.
The thrust magnitude for the direct trajectory in Fig. \ref{fig:thrust_flat}  and the detour trajectory in Fig. \ref{fig:thrust_escape} show that the thrust is indeed lower when allowing detour trajectories by allowing larger TOFs. This is confirmed by the experienced gravity force as a function of time shown in Fig. \ref{fig:gravity_flat} and \ref{fig:gravity_escape}, respectively, where the three centers of gravity are increasingly experienced as one center of gravity at increasing distance. 

We conclude that (i) transformed PINNs can indeed find physically meaningful trajectories according to their constraints and that (ii) the TOF has a major influence on the experienced gravitational force and hence on the loss landscape and in turn on the convergence behavior. The transition between the two regimes may indicate that the bifurcation causes a high-loss barrier between the two local optima that result in the two qualitatively distinct family of trajectories. In other words, if the PINN is initialized in either one regime, a high-loss barrier prevents a gradient-based optimizer from ever exploring the other regime. Whilst our pre-condition step from Sec. \ref{sec:example2} can mitigate this issue, it remains future work to obtain full control of convergence behavior e.g. by finding appropriate annealing schedules for the TOF. Nevertheless, the fact that the TOF can be treated as an auto-differentiable optimization variable is a major advantage of the PINN framework over traditional numerical schemes, which is exacerbated in that the TOF in trajectory optimization usually appears in the boundary conditions.

\section*{Funding Sources}

S.R. was financially supported by the Austrian Science Fund (FWF), Grant DOI: 10.55776/J4774.

\section*{Acknowledgments}
LLMs were used for stylistic improvements of the introduction.

\bibliography{references}

\end{document}